\PassOptionsToPackage{unicode}{hyperref}
\PassOptionsToPackage{hyphens}{url}
\documentclass[
]{article}
\usepackage{amsmath,amssymb}
\usepackage{iftex}
\ifPDFTeX
  \usepackage[T1]{fontenc}
  \usepackage[utf8]{inputenc}
  \usepackage{textcomp} % provide euro and other symbols
\else % if luatex or xetex
  \usepackage{unicode-math} % this also loads fontspec
  \defaultfontfeatures{Scale=MatchLowercase}
  \defaultfontfeatures[\rmfamily]{Ligatures=TeX,Scale=1}
\fi
\usepackage{lmodern}
\ifPDFTeX\else
\fi
\IfFileExists{upquote.sty}{\usepackage{upquote}}{}
\IfFileExists{microtype.sty}{% use microtype if available
  \usepackage[]{microtype}
  \UseMicrotypeSet[protrusion]{basicmath} % disable protrusion for tt fonts
}{}
\makeatletter
\@ifundefined{KOMAClassName}{% if non-KOMA class
  \IfFileExists{parskip.sty}{%
    \usepackage{parskip}
  }{% else
    \setlength{\parindent}{0pt}
    \setlength{\parskip}{6pt plus 2pt minus 1pt}}
}{% if KOMA class
  \KOMAoptions{parskip=half}}
\makeatother
\usepackage{xcolor}
\usepackage[margin=1in]{geometry}
\usepackage{longtable,booktabs,array}
\usepackage{calc} % for calculating minipage widths
\usepackage{etoolbox}
\makeatletter
\patchcmd\longtable{\par}{\if@noskipsec\mbox{}\fi\par}{}{}
\makeatother
\IfFileExists{footnotehyper.sty}{\usepackage{footnotehyper}}{\usepackage{footnote}}
\makesavenoteenv{longtable}
\usepackage{graphicx}
\makeatletter
\def\maxwidth{\ifdim\Gin@nat@width>\linewidth\linewidth\else\Gin@nat@width\fi}
\def\maxheight{\ifdim\Gin@nat@height>\textheight\textheight\else\Gin@nat@height\fi}
\makeatother
\setkeys{Gin}{width=\maxwidth,height=\maxheight,keepaspectratio}
\makeatletter
\def\fps@figure{htbp}
\makeatother
\NewDocumentCommand\citeproctext{}{}

\makeatletter
 \let\@cite@ofmt\@firstofone
 \def\@biblabel#1{}
 \def\@cite#1#2{{#1\if@tempswa , #2\fi}}
\makeatother
\newlength{\cslhangindent}
\newlength{\csllabelwidth}
\newenvironment{CSLReferences}[2] % #1 hanging-indent, #2 entry-spacing
 {\begin{list}{}{%
  \setlength{\itemindent}{0pt}
  \setlength{\leftmargin}{0pt}
  \setlength{\parsep}{0pt}
  \ifodd #1
   \setlength{\leftmargin}{\cslhangindent}
   \setlength{\itemindent}{-1\cslhangindent}
  \fi
  \setlength{\itemsep}{#2\baselineskip}}}
 {\end{list}}
\usepackage{calc}

\ifLuaTeX
  \usepackage{selnolig}  % disable illegal ligatures
\fi
\usepackage{bookmark}
\IfFileExists{xurl.sty}{\usepackage{xurl}}{} % add URL line breaks if available
\hypersetup{
  pdftitle={Generic Embedding-Based Lexicons for Transparent and Reproducible Text Scoring},
  pdfauthor={Catherine Moez},
  hidelinks,
  pdfcreator={LaTeX via pandoc}}

\title{Generic Embedding-Based Lexicons for Transparent and Reproducible
Text Scoring}
\author{Catherine Moez}
\date{November 1, 2024}

\begin{document}
\maketitle

\subsection{Abstract}\label{abstract}

\(\qquad\) With text analysis tools becoming increasingly sophisticated
over the last decade, researchers now face a decision of whether to use
state-of-the-art models that provide high performance but that can be
highly opaque in their operations and computationally intensive to run.
The alternative, frequently, is to rely on older, manually crafted
textual scoring tools that are transparently and easily applied, but can
suffer from limited performance. I present an alternative that combines
the strengths of both: lexicons created with minimal researcher inputs
from generic (pretrained) word embeddings. Presenting a number of
conceptual lexicons produced from FastText and GloVe (6B) vector
representations of words, I argue that embedding-based lexicons respond
to a need for transparent yet high-performance text measuring tools.

\hfill\break

\newpage

\subsection{Introduction}\label{introduction}

\hfill\break

\(\qquad\) Text analysis techniques have become increasingly
sophisticated since the early 2010s turn away from literal text and
toward embedding (numeric vector) representations of words, which are
widely thought to capture semantic meaning. Social science researchers
now have a range of options for high-performance tools, such as large
language models (LLMs), customized supervised learning models, or
pre-defined supervised learning models to score and classify texts. The
main alternative is, for many, applying older, simpler instruments such
as manually crafted conceptual lexicons that may dramatically
underperform state-of-the-art solutions.

\(\qquad\) However, model performance is not the only concern of
interest to many social science researchers. Transparency in which text
features most strongly influence model-generated labels or scores, the
replicable use of the same measuring tool over different domains or
corpuses, ease of use, and the reproducibility of results upon repeated
scoring of the same text with the same measurement instrument are
important to many. Perhaps reflecting these concerns, demand for simple
tools and models remains high: Linguistic Inquiry and Word Count (LIWC)
word lists, first developed in the 1990s, have been referenced in 14,700
scientific articles (per Google Scholar search) since 2020 alone,
despite the emergence of many complex models for sentiment and other
conceptual scoring during this time.

\hfill\break

\paragraph{The Problem}\label{the-problem}

\hfill\break

\(\qquad\) High-performance models are most suitable where accuracy or
other performance metrics are crucial. They do, however, pose a number
of problems to researchers. Highly customized models, such as those
where the researcher generates custom word embeddings (vector space
representations of words) and subsequently measures documents in
relation to these (Rheault et al. 2016; Gennaro and Ash 2022), are not
readily reusable across different research projects as an identical
measuring instrument. They may also suffer from conceptual overfitting
to the corpus under study; words may be associated with emotional or
other connotations that they would not have in general English speech.

\(\qquad\) Out-of-the-box models such as web-trained large language
models (LLMs), likewise, are typically minimally transparent in how they
operate, and can randomly re-score texts differently upon repeated
requests to provide annotations. Although text-generating LLMs are
primarily built around an (embedding-based) infrastructure for next-word
prediction, many of the most popular LLMs are proprietary, black-box
systems where their interior workings are not fully known. Elements such
as safety-focused or anti-bias `guardrails' are known or suspected to
exist in many opaque models (Field 2024), as are attempts to program in
other logics such as math or machine translation, and various degrees of
randomness in token generation (perhaps to avoid direct plagiarism of
training texts). These added elements diverge from a framework of
statistically likely next-word prediction alone.

\(\qquad\) Leading text analysis scholars have concerns about the use of
`closed' proprietary models (Palmer, Smith, and Spirling 2024);
transparency of method is a core value in science, after all.\footnote{Machine
  learning scholars have developed techniques for reverse engineering
  which features matter to outcome scores, such as shuffling independent
  variable columns, but these may be computationally or
  labour-intensive.} On a more practical note, LLM performance quality
can be highly variable (Spirling, Barrie, and Palmer 2024), with
researchers potentially unaware of when it is failing. In addition to
model opaqueness, researchers face steep transaction costs in learning
relevant software languages, ensuring packages and versions are in
order, and obtaining model-training datasets that may no longer be fully
available (as with Twitter-based labelled data). Making modifications to
complex models can likewise be virtually impossible without extensive
experience in the relevant programming languages. Complex models thus
have theoretical downsides relating to transparency, usability,
replicable scoring, and measuring instrument re-usability across
different domains of text.

\(\qquad\) On the other hand, manually created tools such as the
Linguistic Inquiry and Word Count (LIWC) word lists (Boyd et al. 2022),
or the Moral Foundations Dictionary (MFD) created to accompany Jonathan
Haidt and Jesse Graham's moral foundations theory (Graham, Haidt, and
Nosek 2009; Frimer et al. 2019) are readily available and easy to apply
by researchers, but can be small in size and miss relevant terms
(Araque, Gatti, and Kalimeri 2020), or contain creator-specific
particularities, as noted by Aslanidis (2018) in the context of populism
dictionaries. Word denotations and connotations might be suitable to one
domain of text known by the dictionary creator (such as business
documents, or psychological applications) but be inappropriate for
another, such as political speech. The continued widespread use of these
lexicons, however, shows that there is demand for relatively simple,
easily accessible, and transparent tools in text scoring for conceptual
content.

\hfill\break

\paragraph{Generic Embedding Lexicons}\label{generic-embedding-lexicons}

\hfill\break

\(\qquad\) What if, however, the benefits of transparent,
domain-agnostic tools could be combined with some of the gains in
textual analysis made in recent years by the popularization of dense
vector representations of words? I propose a set of generic
(`pretrained', on large, general-purpose corpuses of English text) word
embedding lexicons that draw from the highlights of both approaches.
Requiring relatively minimal researcher inputs to instantiate and
populate a new lexicon, embedding-based lexicons blend the semantic
similarity gains made possible by numeric representations of words with
a fully transparent format for both lexicon creation and text scoring.
The method of creating one's own lexicon is presented, and a number of
concept-specific lexicons are released -- in simple, user-transparent
.csv format -- with this article for researcher use.

\hfill\break

\paragraph{Details of Method}\label{details-of-method}

\hfill\break

\(\qquad\) To create a new conceptual lexicon, the only inputs needed
are a set of `seed words' representing two opposing poles of meaning on
a conceptual dimension, and a set of word embedding numeric vector
representations.\footnote{In theory, even one set of conceptual words
  could be used to identify more with similar meanings, without defining
  an opposite pole; however, all formulas and examples presented use the
  two poles of meaning framework.} The lexicons draw on Rheault et al.
(2016)'s creation of lexicons based on custom word embeddings. I diverge
from Rheault et al. (2016)'s method in the source of the word vector
representations used to populate the rest of the lexicon. As opposed to
customized embeddings, where words might have emotional connotations
that are peculiar to a given corpus under study, I instead use
pretrained FastText and GloVe vector representations of words and use
these as the basis for lexicon expansion.

\(\qquad\) GloVe, in the 300-dimensional, `6B' (six billion word)
trained version I use (Pennington, Socher, and Manning 2014), is
estimated from word co-occurrences in a large corpus made up of a
snapshot of Wikipedia in 2014 and Gigaword, a collection of several
decades of newswire text. The pretrained version of FastText (Joulin et
al. 2016) I use is trained on Common Crawl, a large collection of
web-scraped text. (Please note that GloVe is also available in versions
estimated on Common Crawl text, and on Twitter data). The two
vocabularies are initially 400,000 English words in size (GloVe) or 2
million (FastText); after preliminary editing to remove proper nouns,
excessively common words (top 50 in rank), and mis-spellings, roughly
77,000 candidate terms remain in GloVe and 100,000 in FastText.

\(\qquad\) To expand the dictionary beyond the small initial list of
seed words, the words closest in vector space to the two poles of
meaning are identified, and retained in the new, expanded lexicon. Words
with the highest cosine similarity to the positive pole (the sum of all
pairwise cosine similarities), net of their summed cosine similarities
to words at the negative pole, thus become added `positive' lexicon
words. Those with the highest similarity to the negative pole, likewise,
become the added negative word list, as in the formula reported in
Rheault et al. (2016). The released lexicons include these scores,
ranging from -1 to +1, as `polarity'.

\(\qquad\) The final lexicon size is arbitrary. Researchers creating a
lexicon are encouraged to manually inspect for the point where returned
words appear to be much less relevant to the concept named, and adjust
dictionary size accordingly, especially in applications where each
dictionary word counts equally toward the concept it represents. (In the
Quanteda text analysis package for R, this type of scoring is known as
`polarity' scoring and is the default). Alternatively, if using the
conceptual intensity or extremity scores in scoring (`valence' scoring,
where each word may be weighted differently, in Quanteda), the total
size of the lexicon is of less importance, as added words' similarity
scores will naturally decrease as more are added.

\hfill\break

\paragraph{Comparison to Existing Models and
Lexicons}\label{comparison-to-existing-models-and-lexicons}

\hfill\break

\(\qquad\) Previous iterations of embedding-based lexicons or
embedding-based text scoring have typically required the researcher to
estimate their own custom embeddings from their corpus (Rheault et al.
2016; Gennaro and Ash 2022)\footnote{Gennaro and Ash begin with LIWC
  word lists to construct a larger emotionality dictionary, but then
  scale their corpus (with custom embeddings) and assess texts' semantic
  similarities to their two poles of meaning, rather than applying a
  `word scoring' approach (2022, 1041-1043).}, or to obtain pretrained
embeddings and carry out similarity measurements between the documents
under study and relevant theoretical words (Lin et al. 2018; Araque,
Gatti, and Kalimeri 2020). Custom embedding estimations have the benefit
of linking words to their corpus-specific connotations. As Rheault et
al. (2016) point out, policy discussions of an issue such as `war' or
`health' care should likely not be considered as emotion-conveying as
mentions of these terms in a personal conversation would warrant.
However, the customized embedding approach risks essentially
`overfitting' emotional connotations of words to the corpus used for
measuring tool estimation (and then measured with the same instrument).
Images in the Appendix highlight words that have large discrepancies in
sentiment connotation between the British parliamentary record and
`generic' English text-trained embeddings.

\(\qquad\) Other recent applications of text measurement that draw at
least partially on vector (embedding) representations of words include
Lin et al. (2018), who use pretrained Word2Vec vectors trained on Google
News text (\url{https://code.google.com/archive/p/word2vec/}), and
aggregate these into document embeddings -- an average or other
aggregation of word embedding representations over a longer text -- as
part of a classifier-based approach to identifying moral foundations in
Twitter posts. Further development of embedding-leveraging text scoring
has also been applied by Araque, Zhu, and Iglesias (2019) and Araque,
Gatti, and Kalimeri (2020), who measure the vector similarity of texts
to an existing moral foundations lexicon, rather than using the
embeddings to expand the lexicon itself. This approach, also implemented
in Gennaro and Ash (2022), requires more familiarity with estimating or
obtaining embeddings and taking similarity measurements in vector space,
compared to the conventional word scoring method. It may, furthermore,
produce similar results overall to using a dictionary scoring approach
with an embedding-informed lexicon.

\newpage

\begin{longtable}[]{@{}
  >{\raggedright\arraybackslash}p{(\columnwidth - 8\tabcolsep) * \real{0.1702}}
  >{\raggedright\arraybackslash}p{(\columnwidth - 8\tabcolsep) * \real{0.1418}}
  >{\raggedright\arraybackslash}p{(\columnwidth - 8\tabcolsep) * \real{0.0567}}
  >{\raggedright\arraybackslash}p{(\columnwidth - 8\tabcolsep) * \real{0.3014}}
  >{\raggedright\arraybackslash}p{(\columnwidth - 8\tabcolsep) * \real{0.3298}}@{}}
\caption{Comparison of Sentiment and Emotionality
Lexicons}\tabularnewline
\toprule\noalign{}
\begin{minipage}[b]{\linewidth}\raggedright
Lexicon
\end{minipage} & \begin{minipage}[b]{\linewidth}\raggedright
Size
\end{minipage} & \begin{minipage}[b]{\linewidth}\raggedright
Date
\end{minipage} & \begin{minipage}[b]{\linewidth}\raggedright
WordSelection
\end{minipage} & \begin{minipage}[b]{\linewidth}\raggedright
WordValidation
\end{minipage} \\
\midrule\noalign{}
\endfirsthead
\toprule\noalign{}
\begin{minipage}[b]{\linewidth}\raggedright
Lexicon
\end{minipage} & \begin{minipage}[b]{\linewidth}\raggedright
Size
\end{minipage} & \begin{minipage}[b]{\linewidth}\raggedright
Date
\end{minipage} & \begin{minipage}[b]{\linewidth}\raggedright
WordSelection
\end{minipage} & \begin{minipage}[b]{\linewidth}\raggedright
WordValidation
\end{minipage} \\
\midrule\noalign{}
\endhead
\bottomrule\noalign{}
\endlastfoot
Linguistic Inquiry and Word Count (LIWC) & 12,000 (LIWC-22) & Early-mid
1990s & Hand selected (Theoretically informed) & Expert judgement, now
supplemented with statistical measures \\
NRC Emotion Lexicon & 141,539 (v. 0.92) & 2011 & Random (Initiated from
large English word lists) & Crowdsourced via Mechanical Turk \\
Lexicoder Sentiment Dictionary & 4,159 (v. 3.0) & 2012 & Merged from
three prior sentiment word lists & `Three trained human coders'
classifying sentiment of 900 news articles \\
PLOSlex (Rheault et al.~2016) & 4,200 & 2016 & Hand-selected seed words
+ Nearest embeddings in a custom vector space & Face validation (p.4)
and tests against available sentiment datasets (Appendix) \\
Emotion vs.~Cognition (Gennaro and Ash 2022) & 1,189 & 2022 & Drawn from
LIWC & Face validation and human annotators assessing `thousands' of
ranked sentence pairs (p.1038) \\
Generic Embedding Lexicons (FastText and GloVe) & 3,000 (Sentiment);
3,400 (Emotionality) & 2024 & Seed words from prior publications +
Nearest embeddings in a pretrained vector space & Face validation and
tests against available sentiment datasets \\
\end{longtable}

\begin{longtable}[]{@{}
  >{\raggedright\arraybackslash}p{(\columnwidth - 8\tabcolsep) * \real{0.2087}}
  >{\raggedright\arraybackslash}p{(\columnwidth - 8\tabcolsep) * \real{0.1087}}
  >{\raggedright\arraybackslash}p{(\columnwidth - 8\tabcolsep) * \real{0.0217}}
  >{\raggedright\arraybackslash}p{(\columnwidth - 8\tabcolsep) * \real{0.4261}}
  >{\raggedright\arraybackslash}p{(\columnwidth - 8\tabcolsep) * \real{0.2348}}@{}}
\caption{Comparison of Moral Foundations Lexicons}\tabularnewline
\toprule\noalign{}
\begin{minipage}[b]{\linewidth}\raggedright
Lexicon
\end{minipage} & \begin{minipage}[b]{\linewidth}\raggedright
Size
\end{minipage} & \begin{minipage}[b]{\linewidth}\raggedright
Date
\end{minipage} & \begin{minipage}[b]{\linewidth}\raggedright
WordSelection
\end{minipage} & \begin{minipage}[b]{\linewidth}\raggedright
WordValidation
\end{minipage} \\
\midrule\noalign{}
\endfirsthead
\toprule\noalign{}
\begin{minipage}[b]{\linewidth}\raggedright
Lexicon
\end{minipage} & \begin{minipage}[b]{\linewidth}\raggedright
Size
\end{minipage} & \begin{minipage}[b]{\linewidth}\raggedright
Date
\end{minipage} & \begin{minipage}[b]{\linewidth}\raggedright
WordSelection
\end{minipage} & \begin{minipage}[b]{\linewidth}\raggedright
WordValidation
\end{minipage} \\
\midrule\noalign{}
\endhead
\bottomrule\noalign{}
\endlastfoot
MFD (2009/2019) & 4,808 (v. 2; 324 in v.1) & 2009 & Selected from LIWC &
Theoretical (LIWC) \\
MoralStrength (Araque et al.~2020) & 1,000 & 2020 & Words in MFD v.1,
expanded with WordNet synsets & Crowdsourced (5 annotators x 1,000
words) \\
EMFD (Hopp et al.~2021) & 3,270 & 2021 & Crowdsourced from annotations
of news texts & Crowdsourced (854 annotators x 15 news texts each) \\
MoralBERT & Not lexicon-based & 2024 & Not lexicon-based; trained on
labelled data & Performance against labelled data \\
Generic Embedding Lexicons (FastText and GloVe) & 10,000 & 2024 & Seed
words from prior publications and lexicons + Nearest embeddings in a
pretrained vector space & Face validation and performance against
labelled data \\
\end{longtable}

\hfill\break

\paragraph{Outline}\label{outline}

\hfill\break

\(\qquad\) Below, I first present a `face validation' of whether the
words returned to populate the lexicon, automatically (by a cosine
similarity measurement, in effect), are `intuitive and satisfying', as
in Gennaro and Ash (2022). Full lists of the most extreme 5-10 words
returned at each pole are presented in the Appendix. (Images show only a
sampling of words from each lexicon, for labelling purposes). Next,
where annotated datasets are available, I report conventional accuracy
and F1 scores for categorical labels, as well as regression performance
metrics where outcomes are continuous. For lexicons where annotated
datasets are not readily available, I report examples of extreme-scoring
text excerpts for distinct positive emotions such as hope or nostalgia.

\(\qquad\) As a reminder, the aim of the generic embedding lexicons is
not necessarily the highest performance on all datasets. More complex
models, such as MoralBERT (Preniqi et al. 2024) for moral frames
identification, or LLMs based on extensive training corpuses and complex
embedding representations, like Chat GPT, are expected to exceed the
relatively simple embedding-based lexicons presented. However, if
performance gaps are not substantially large, or if transparency is of
high importance, generic embedding lexicons can nonetheless provide a
suitable alternative for conceptual measurement in texts.

\(\qquad\) Lexicons based on intensive inputs of human time -- either in
theoretically generating word lists, as for LIWC and the MFD, or in the
extensive use of trained and untrained coders to annotate words with
meaning -- are also likely to perform well. Generic embedding lexicons
nonetheless serve a purpose in that they are adaptable to capture
virtually any conceptual dimension a researcher wishes to measure,
without requiring the resources that hiring annotators for thousands of
word comparisons, as in Saif M. Mohammad and Turney (2013), Young and
Soroka (2012), Araque, Gatti, and Kalimeri (2020), or Hopp et al.
(2021).

\hfill\break

\subsection{Generic Embedding
Dictionaries}\label{generic-embedding-dictionaries}

\hfill\break

\subsubsection{Sentiment and
Emotionality}\label{sentiment-and-emotionality}

\(\qquad\) An initial check of the most extreme words returned at each
pole suggests that the method correctly isolates words similar in
meaning to those inputted as seeds. FastText and GloVe lexicons are
presented separately throughout. Likely due to differences in training
data, FastText includes more informal and web-based language; whereas
GloVe, trained on the news-based Gigaword corpus and Wikipedia, has a
heavier concentration of formal, political or otherwise news-related,
and obscure scientific terms.

\begin{figure}
\centering
\includegraphics{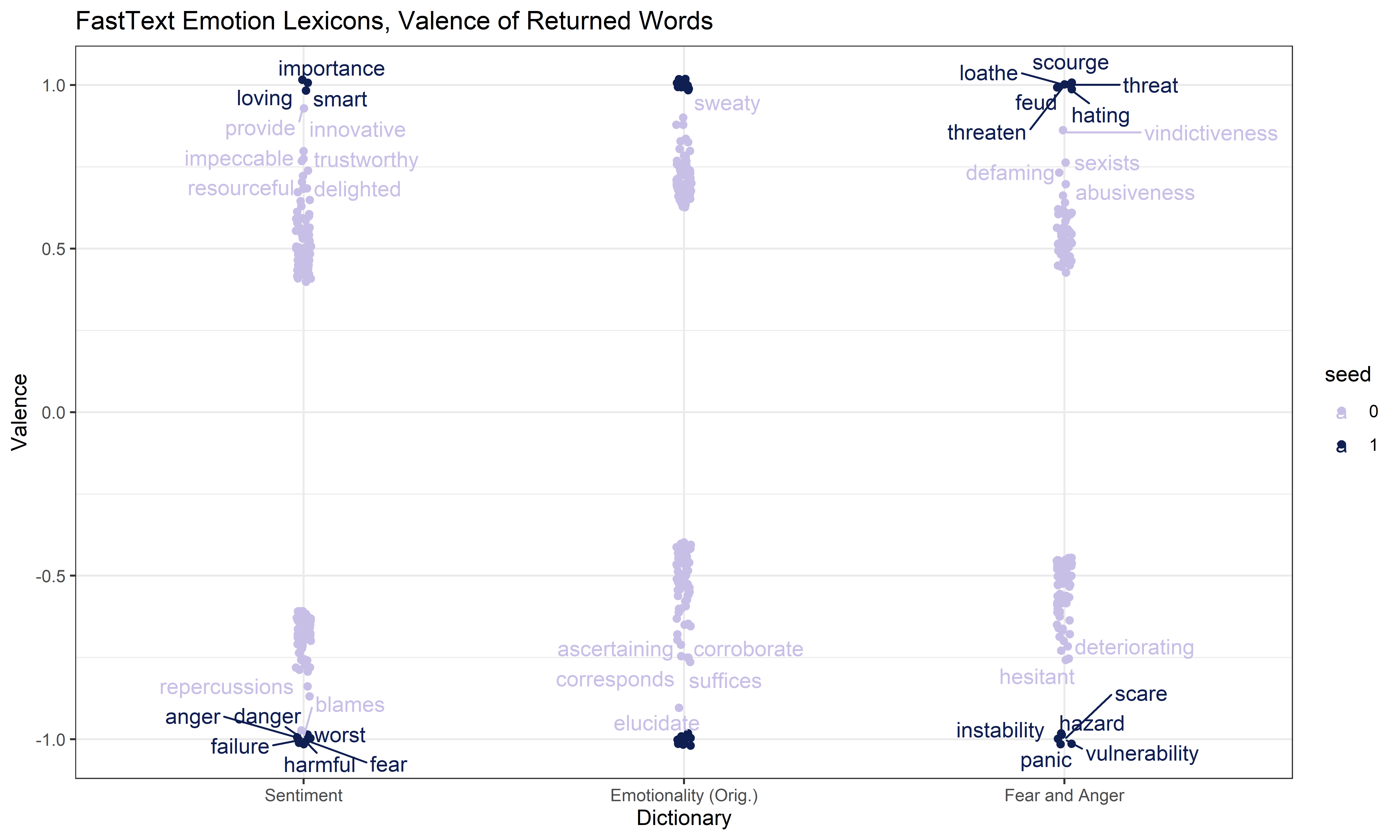}
\caption{FastText embedding-based dictionaries}
\end{figure}

\begin{figure}
\centering
\includegraphics{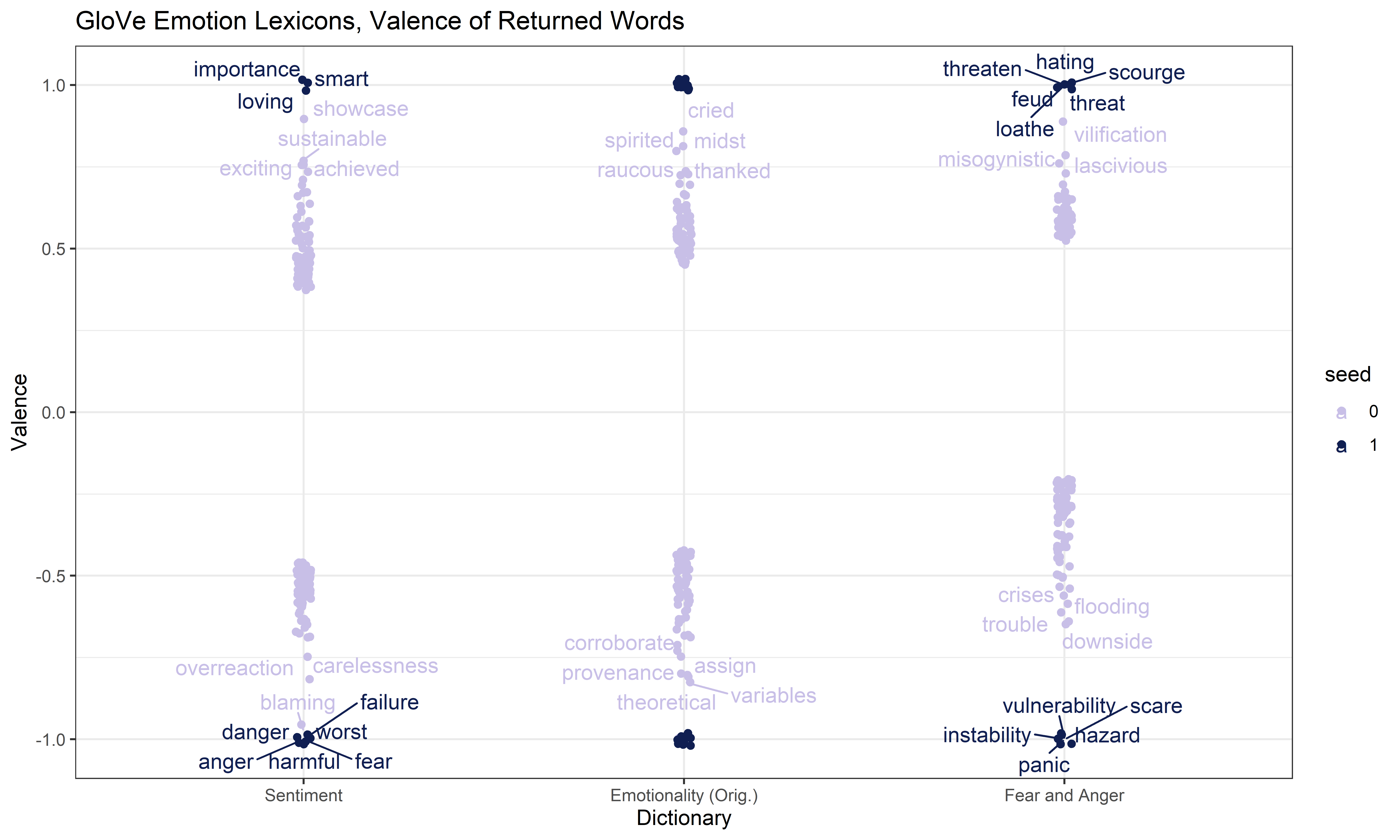}
\caption{GloVe embedding-based dictionaries}
\end{figure}

\subsubsection{Accuracy Tests against Labelled
Data}\label{accuracy-tests-against-labelled-data}

\(\qquad\) GloVe and FastText sentiment lexicons are next compared
against three datasets with sentiment labelled: Amazon Reviews (2023
version, Hou et al. (2024), n = 3000), for a test in a more informal
language setting; Keith and Meneses' Paper Reviews (-Keith and Meneses
(2017), n = 400); and the Financial Phrasebank, a collection of
sentence-long financial news updates (Malo et al. (2014), n = 4,700).
The datasets vary in length from roughly one sentence to a paragraph or
more for Amazon Reviews and the conference paper reviews. Additionally,
they vary in formality, with the Amazon comments taking the form of
informal web speech and the other two datasets a more formal register of
academic reviews and financial news reporting, respectively.\footnote{The
  Spanish language paper reviews are translated to English using the
  October 2024 version of Google Translate. The Amazon reviews consist
  of random samples of 1000 from the Digital Music and Magazine
  Subscriptions categories, and a random sample of 1000 from the last
  100,000 JSON entries for the Books category, due to its large file
  size. They cover from 1997 to 2023 in time.}

\(\qquad\) Comparison sentiment lexicons include the NRC Emotion Lexicon
(Saif M. Mohammad and Turney 2013), Young and Soroka's Lexicoder (Young
and Soroka 2012), and the custom embedding-based dictionary produced by
Rheault et al. (2016), labelled the `PLOS' lexicon or `PLOSlex' below.
Alternative dictionaries might include Linguistic Inquiry and Word Count
(LIWC), which is proprietary software and is not included here in tests.
More `advanced' sentiment scoring is provided by LLM output from Chat
GPT 4o, produced over 18-19 October 2024. (GPT 4o reports that it
applies the TextBlob library for sentiment scoring, and that this
approach is based not only on unigram (1-word) or bag-of-words
(unordered) word usage, but also on multiple-word parsing, as well as
the `consistency' of negative or positive language in a text).
Alternative models and lexicons might also include VADER and sentimentr.

\hfill\break

\paragraph{Scoring Details}\label{scoring-details}

\hfill\break

\(\qquad\) For categorical outcomes, accuracy and F1 scores are
reported:

\[Accuracy = \frac{(TP + TN)}{n}\]

\[Precision = \frac{TP}{(TP + FP)}\]

\[Recall = \frac{TP}{(TP + FN)}\]

\[F1 = 2\times\frac{Precision \times Recall}{Precision + Recall}\]

where \(TP\) = true positive, \(TN\) = true negative, \(FP\) = false
positive, \(FN\) = false negative, and \(n\) is the total number of
observations.\footnote{For multi-class outcomes, as in the prediction of
  moral frames, each frame's F1 score is computed separately, then all
  are averaged}.

\(\qquad\) Accuracy scores can give a false impression of good
performance when outcome classes are highly imbalanced (e.g., predicting
all scores to the most common class can give high Accuracy scores); F1
scores address this concern by balancing a penalization of false
positives (Precision) with a concern for false negatives (Recall).

\(\qquad\) For continuous outcomes, root mean square error (RMSE), in
effect the average error size, is reported, as is adjusted R-Squared as
a measure of the proportion of variance explained by the model. The RMSE
formula follows,

\[RMSE = \sqrt \frac{\sum (x_i - \hat x_i)^2}{ N }\]

where x is the actual outcome values, \(\hat x\) is their model-based
predictions, and N is the valid number of cases.

\hfill\break

\paragraph{Paper Reviews dataset}\label{paper-reviews-dataset}

\hfill\break

\(\qquad\) For the conference paper reviews dataset, models are ranked
from best (smallest RMSE) to worst over two alternative linear
regressions. The first set of models' uses evaluators' scores as the
dependent variable (DV); the next test uses the alternative DV of the
researchers' (Keith and Meneses 2017) subjective assessments of the
ratings' positivity. The sole independent variable in each model is the
text-based net positivity score.

\begin{longtable}[]{@{}
  >{\raggedright\arraybackslash}p{(\columnwidth - 6\tabcolsep) * \real{0.1667}}
  >{\raggedleft\arraybackslash}p{(\columnwidth - 6\tabcolsep) * \real{0.1667}}
  >{\raggedleft\arraybackslash}p{(\columnwidth - 6\tabcolsep) * \real{0.1250}}
  >{\raggedright\arraybackslash}p{(\columnwidth - 6\tabcolsep) * \real{0.5417}}@{}}
\caption{Performance of FastText and GloVe Sentiment Lexicons on
Conference Papers Dataset}\tabularnewline
\toprule\noalign{}
\begin{minipage}[b]{\linewidth}\raggedright
Lexicon
\end{minipage} & \begin{minipage}[b]{\linewidth}\raggedleft
AdjRSquared
\end{minipage} & \begin{minipage}[b]{\linewidth}\raggedleft
RMSE
\end{minipage} & \begin{minipage}[b]{\linewidth}\raggedright
Test
\end{minipage} \\
\midrule\noalign{}
\endfirsthead
\toprule\noalign{}
\begin{minipage}[b]{\linewidth}\raggedright
Lexicon
\end{minipage} & \begin{minipage}[b]{\linewidth}\raggedleft
AdjRSquared
\end{minipage} & \begin{minipage}[b]{\linewidth}\raggedleft
RMSE
\end{minipage} & \begin{minipage}[b]{\linewidth}\raggedright
Test
\end{minipage} \\
\midrule\noalign{}
\endhead
\bottomrule\noalign{}
\endlastfoot
Lexicoder & 0.084676 & 1.440087 & Conference Papers, Evaluators'
Ratings \\
GPT 4o & 0.041529 & 1.473638 & Conference Papers, Evaluators' Ratings \\
FastText & 0.039831 & 1.474943 & Conference Papers, Evaluators'
Ratings \\
PLOSlex & 0.034361 & 1.479139 & Conference Papers, Evaluators'
Ratings \\
GloVe & 0.021934 & 1.488626 & Conference Papers, Evaluators' Ratings \\
NRC Emotion & 0.018154 & 1.491499 & Conference Papers, Evaluators'
Ratings \\
\end{longtable}

\begin{longtable}[]{@{}
  >{\raggedright\arraybackslash}p{(\columnwidth - 8\tabcolsep) * \real{0.0395}}
  >{\raggedright\arraybackslash}p{(\columnwidth - 8\tabcolsep) * \real{0.1579}}
  >{\raggedleft\arraybackslash}p{(\columnwidth - 8\tabcolsep) * \real{0.1579}}
  >{\raggedleft\arraybackslash}p{(\columnwidth - 8\tabcolsep) * \real{0.1184}}
  >{\raggedright\arraybackslash}p{(\columnwidth - 8\tabcolsep) * \real{0.5263}}@{}}
\caption{Performance of FastText and GloVe Sentiment Lexicons on
Conference Papers Dataset}\tabularnewline
\toprule\noalign{}
\begin{minipage}[b]{\linewidth}\raggedright
\end{minipage} & \begin{minipage}[b]{\linewidth}\raggedright
Lexicon
\end{minipage} & \begin{minipage}[b]{\linewidth}\raggedleft
AdjRSquared
\end{minipage} & \begin{minipage}[b]{\linewidth}\raggedleft
RMSE
\end{minipage} & \begin{minipage}[b]{\linewidth}\raggedright
Test
\end{minipage} \\
\midrule\noalign{}
\endfirsthead
\toprule\noalign{}
\begin{minipage}[b]{\linewidth}\raggedright
\end{minipage} & \begin{minipage}[b]{\linewidth}\raggedright
Lexicon
\end{minipage} & \begin{minipage}[b]{\linewidth}\raggedleft
AdjRSquared
\end{minipage} & \begin{minipage}[b]{\linewidth}\raggedleft
RMSE
\end{minipage} & \begin{minipage}[b]{\linewidth}\raggedright
Test
\end{minipage} \\
\midrule\noalign{}
\endhead
\bottomrule\noalign{}
\endlastfoot
7 & FastText & 0.109256 & 0.968907 & Conference Papers, Researchers'
Ratings \\
8 & Lexicoder & 0.107597 & 0.969809 & Conference Papers, Researchers'
Ratings \\
9 & GPT 4o & 0.085071 & 0.981973 & Conference Papers, Researchers'
Ratings \\
10 & PLOSlex & 0.065223 & 0.992566 & Conference Papers, Researchers'
Ratings \\
11 & GloVe & 0.056455 & 0.997211 & Conference Papers, Researchers'
Ratings \\
12 & NRC Emotion & 0.027884 & 1.012196 & Conference Papers, Researchers'
Ratings \\
\end{longtable}

\paragraph{Amazon Reviews dataset}\label{amazon-reviews-dataset}

\hfill\break

\(\qquad\) Models' performance against Amazon book, digital music, and
magazine reviews over 1996 to 2023 (Hou et al. 2024) are next reported.
Review ratings range from 0 to 5; please note that the Books reviews are
almost universally divided between 0s and 5s, while Digital Music and
Magazine Subscriptions reviews consist of mainly 5 out of 5 scores.

As before, the best models by smallest RMSE are ranked for each test.
Additional models where customer ratings are split into two (Positive,
Negative) or three (Positive, Neutral, Negative) categories are shown as
logistic regressions in the Appendix.

\begin{longtable}[]{@{}
  >{\raggedright\arraybackslash}p{(\columnwidth - 8\tabcolsep) * \real{0.0400}}
  >{\raggedright\arraybackslash}p{(\columnwidth - 8\tabcolsep) * \real{0.1600}}
  >{\raggedleft\arraybackslash}p{(\columnwidth - 8\tabcolsep) * \real{0.1600}}
  >{\raggedleft\arraybackslash}p{(\columnwidth - 8\tabcolsep) * \real{0.1200}}
  >{\raggedright\arraybackslash}p{(\columnwidth - 8\tabcolsep) * \real{0.5200}}@{}}
\caption{Performance of FastText and GloVe Sentiment Lexicons on Amazon
Reviews (Linear Regression Models)}\tabularnewline
\toprule\noalign{}
\begin{minipage}[b]{\linewidth}\raggedright
\end{minipage} & \begin{minipage}[b]{\linewidth}\raggedright
Lexicon
\end{minipage} & \begin{minipage}[b]{\linewidth}\raggedleft
AdjRSquared
\end{minipage} & \begin{minipage}[b]{\linewidth}\raggedleft
RMSE
\end{minipage} & \begin{minipage}[b]{\linewidth}\raggedright
Test
\end{minipage} \\
\midrule\noalign{}
\endfirsthead
\toprule\noalign{}
\begin{minipage}[b]{\linewidth}\raggedright
\end{minipage} & \begin{minipage}[b]{\linewidth}\raggedright
Lexicon
\end{minipage} & \begin{minipage}[b]{\linewidth}\raggedleft
AdjRSquared
\end{minipage} & \begin{minipage}[b]{\linewidth}\raggedleft
RMSE
\end{minipage} & \begin{minipage}[b]{\linewidth}\raggedright
Test
\end{minipage} \\
\midrule\noalign{}
\endhead
\bottomrule\noalign{}
\endlastfoot
5 & NRC Emotion & -0.0001282 & 2.364396 & Amazon Reviews (Avge. of 3
Categories) \\
3 & GPT 4o & 0.0009975 & 2.363065 & Amazon Reviews (Avge. of 3
Categories) \\
1 & FastText & 0.0022108 & 2.361630 & Amazon Reviews (Avge. of 3
Categories) \\
2 & GloVe & 0.0028141 & 2.360916 & Amazon Reviews (Avge. of 3
Categories) \\
6 & PLOSlex & 0.0037364 & 2.359824 & Amazon Reviews (Avge. of 3
Categories) \\
4 & Lexicoder & 0.0042634 & 2.359200 & Amazon Reviews (Avge. of 3
Categories) \\
\end{longtable}

\paragraph{Financial Phrase dataset}\label{financial-phrase-dataset}

\hfill\break

\(\qquad\) In the final test of sentiment scoring performance, another
more formal and social-science-relevant dataset is tested: the Financial
Phrasebook (Malo et al. 2014). 16 coders assign categories of
`positive', `neutral', or `negative' to sentence-long lines of financial
news. All entries with at least 50\% coder agreement are retained.

\begin{longtable}[]{@{}llrl@{}}
\caption{Performance of FastText and GloVe Sentiment Lexicons on
Financial Phrasebook Data (Logistic Regression Models)}\tabularnewline
\toprule\noalign{}
Lexicon & Model & Accuracy & Test \\
\midrule\noalign{}
\endfirsthead
\toprule\noalign{}
Lexicon & Model & Accuracy & Test \\
\midrule\noalign{}
\endhead
\bottomrule\noalign{}
\endlastfoot
GloVe & Logistic & 0.598522 & Financial Phrase Bank (3-Class DV) \\
FastText & Logistic & 0.596410 & Financial Phrase Bank (3-Class DV) \\
PLOSlex & Logistic & 0.595776 & Financial Phrase Bank (3-Class DV) \\
Lexicoder & Logistic & 0.595565 & Financial Phrase Bank (3-Class DV) \\
NRC Emotion & Logistic & 0.594509 & Financial Phrase Bank (3-Class
DV) \\
GPT 4o & Logistic & 0.494192 & Financial Phrase Bank (3-Class DV) \\
\end{longtable}

\(\qquad\) In all three datasets, GloVe and FastText models perform
similarly, and in some cases better, than both an LLM model (GPT 4o) and
well-validated lexicons that required extensive manual annotation to
produce (Lexicoder; NRC Emotion Lexicon).

\(\qquad\) In well-developed text-scoring areas, such as the detection
of general positive--negative sentiment in texts, researchers may well
wish to use other options. However, a core benefit of the generic
embedding dictionaries is that they can be generated for concepts that
existing tools do not cover, such as distinct emotions. These would be
prohibitively expensive for many research teams to manually annotate or
manually validate at scale; promisingly, the tests where comparison
lexicons are available show that the low-resource embedding-based
approach can meet or exceed current standards.

\hfill\break

\subsubsection{The Moral Frames Lexicon}\label{the-moral-frames-lexicon}

\hfill\break

\(\qquad\) Moral foundations lexicons produced from FastText and GloVe
pretrained embeddings are presented next. Like sentiment scoring, a
number of competing lexicons and non-word-scoring models exist in this
area, as do `ground truth' labelled datasets to test performance
against. The five primary moral foundations are named after their
`virtue-vice' opposites of Care--Harm, Fairness--Cheating,
Loyalty--Betrayal, Authority--Subversion, and Sanctity (or Purity) --
Degradation (Graham, Haidt, and Nosek 2009; Clifford et al. 2015; Shahid
et al. 2020).

\(\qquad\) Haidt and Graham introduce a 324-token dictionary to
accompany their social-psychological theory in 2009; since this time,
the original MFD has been expanded to over 2,000 words (Frimer et al.
2019), and others have proposed both word overlap (between multiple
frames) and (objective, proportion of word appearances attributed a
moral frame) valence attributions to each word, in the `Extended' Moral
Foundations Dictionary, the EMFD (Hopp et al. 2021). Recent
contributions include Araque, Gatti, and Kalimeri (2020), who expand the
original MFD with more words and the assistance of subjective human
annotations of valence; and Preniqi et al. (2024), who train supervised
machine learning models on Twitter, Facebook, and Reddit labelled data,
with the assistance of BERT embedding representations; they release the
defined model as Python code and an app, available at
(\url{https://huggingface.co/spaces/vjosap/MoralBERTApp}).

\(\qquad\) The GloVe and FastText lexicons compared against these models
contain 2,000 words per frame, 1,000 at each of the `vice' and `virtue'
poles. This exceeds previous lexicons in size; their compared
performance appears below. Seed words for the moral frames lexicons are
again drawn from prior theory and lexicon resources. To illustrate the
advantages of locating closely related words to a concept through a
semantic similarity approach, clearly theoretically relevant conceptual
words such as `cesspit' (for the opposite of Purity) appear for the
first time in the generic embedding dictionaries presented, and do not
appear in the MFD, EMFD, or Araque word lists.

\begin{figure}
\centering
\includegraphics{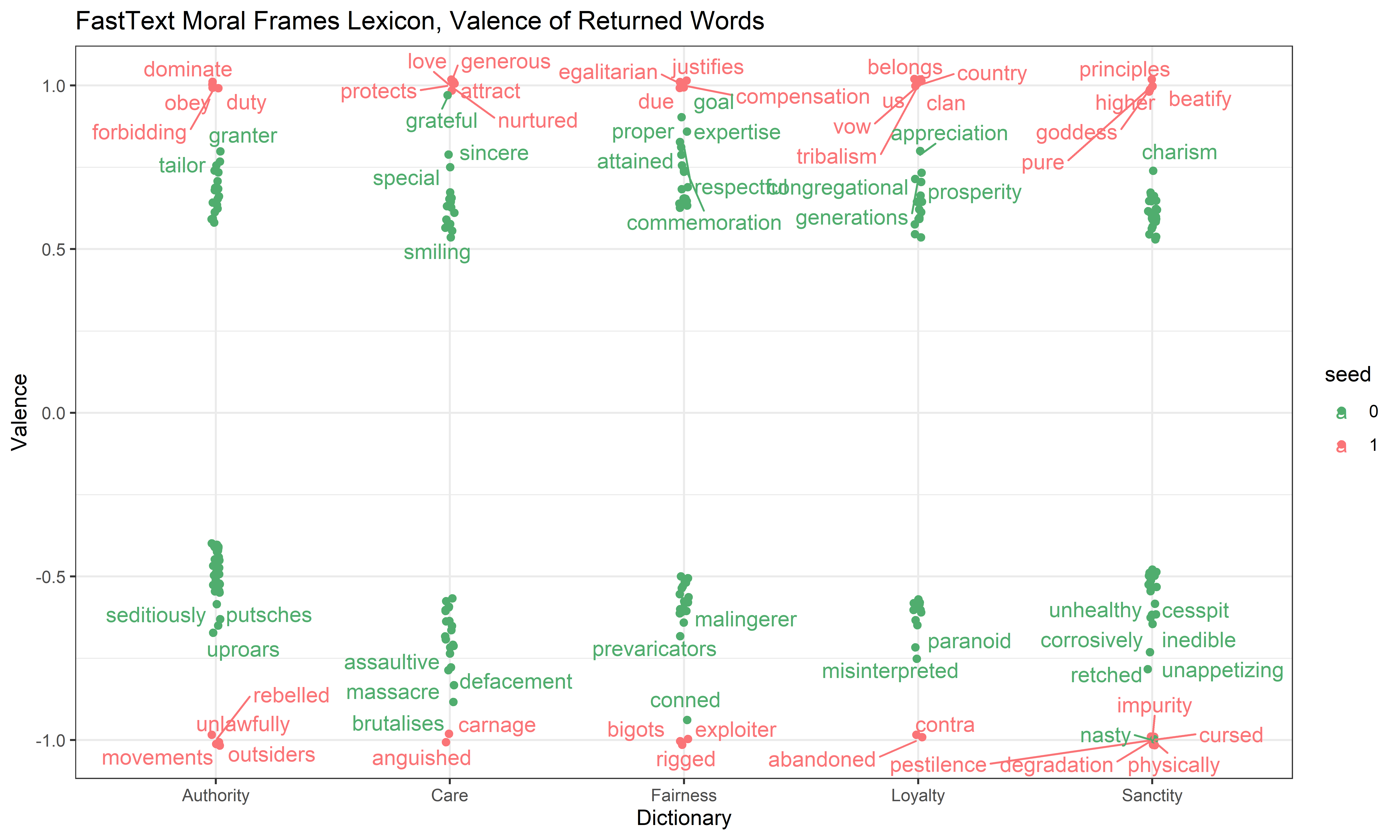}
\caption{FastText embedding-based dictionaries, moral frames lexicon}
\end{figure}

\begin{figure}
\centering
\includegraphics{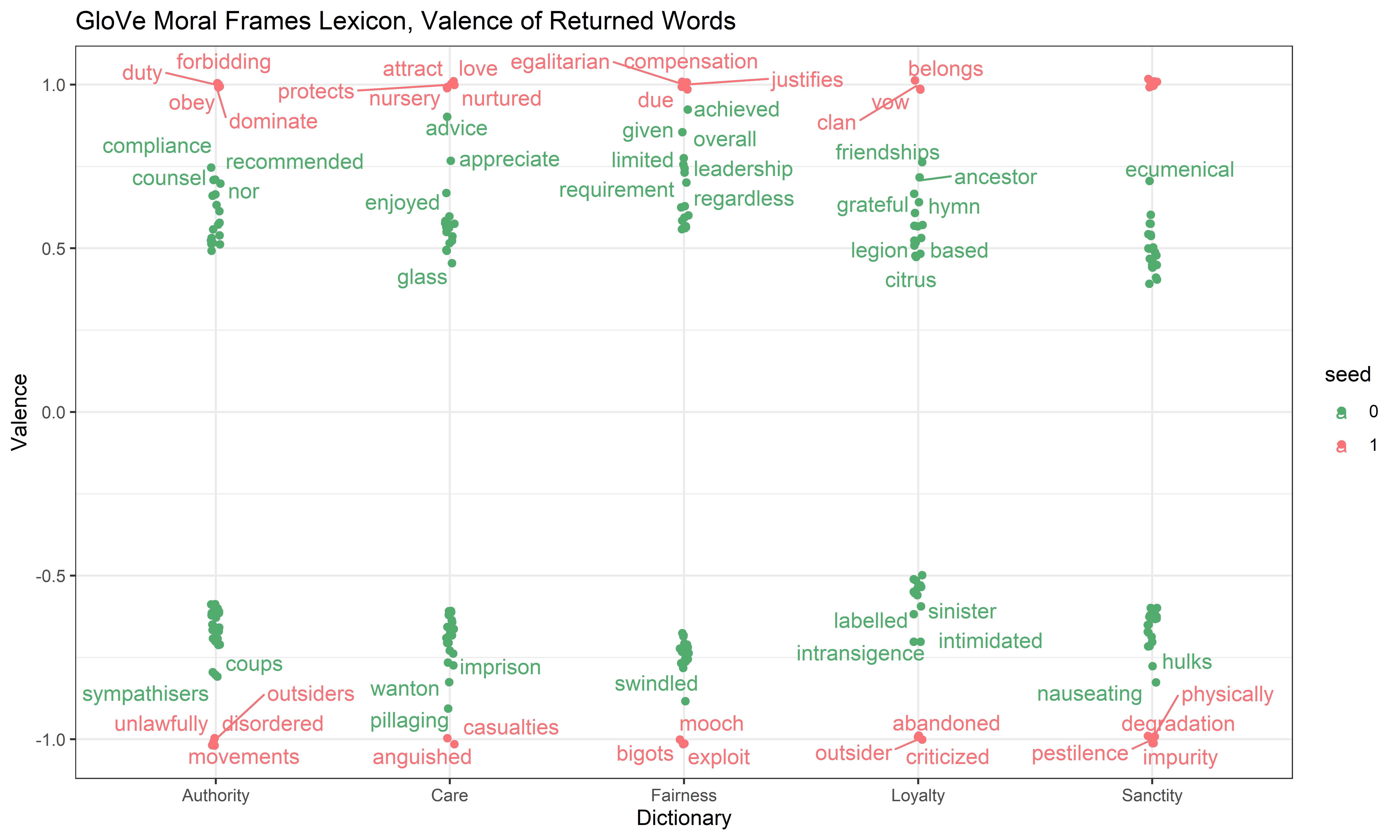}
\caption{GloVe embedding-based dictionaries, moral frames lexicon}
\end{figure}

\(\qquad\) As in other examples, the GloVe lexicon's tendency toward
capturing rarer and more formal words -- such as medical terms -- is
shown.

\hfill\break

\subsubsection{Performance of the Moral Foundations
Lexicons}\label{performance-of-the-moral-foundations-lexicons}

\hfill\break

\(\qquad\) The labelled datasets against which the generic embedding
lexicons and other lexicons and models are tested are the Clifford
collection of moral foundations `vignettes' (2015); three Reddit forums
covering everyday scenarios, French politics discussion, and U.S.
politics (Trager et al. 2022); and an annotated collection of 400 news
articles, where coders label moral frames at the complete document and
line level (Shahid et al. 2020). As in the sentiment tests, these
corpuses range in length from sentence-long texts in Clifford et al.
(2015) to social media posts and full news articles ranging from one to
a few paragraphs in length. They similarly range in formality from
casual online speech to plain English descriptions of social scenarios
and news reporting.

\(\qquad\) Comparator lexicons are the MFD 2.0, EMFD, and two versions
of the Araque, Gatti, and Kalimeri (2020) lexicon available online
(\url{https://github.com/oaraque/moral-foundations}). Additionally, Chat
GPT 4o is again given a version of the data with the `ground truth'
label removed, and asked in a zero-shot manner (i.e., based on the
model's training inputs only) to provide moral frames scores and to name
the most dominant moral frame, if one is present. (The Reddit forums
data includes many posts labelled as `non-moral', whereas the Clifford
and Shahid datasets contain only documents labelled as having moral
content of some kind). The most advanced model, MoralBERT (Preniqi et
al. 2024), draws on a mix of BERT embedding representations of words and
supervised learning. It is expected to perform well on the Reddit data
in particular, given that this is one of the datasets the model was
trained on.

\(\qquad\) For the scoring below, the frame with the highest scores
(summed scores for each frame, regardless of `virtue' or `vice'
polarity; polarity scoring applied) is reported as the model's
prediction. If all frames have a 0 score, the model's prediction is
reported as `Non-moral'. Confusion matrices with the `ground truth'
label are then used to report accuracy and F1 scores. As this is a
multi-class prediction task, F1 scores are obtained for each of the five
moral frames, then averaged to produce one overall F1 score across
categories.

\paragraph{Clifford Vignettes}\label{clifford-vignettes}

\begin{longtable}[]{@{}
  >{\raggedright\arraybackslash}p{(\columnwidth - 6\tabcolsep) * \real{0.2235}}
  >{\raggedleft\arraybackslash}p{(\columnwidth - 6\tabcolsep) * \real{0.1059}}
  >{\raggedleft\arraybackslash}p{(\columnwidth - 6\tabcolsep) * \real{0.1059}}
  >{\raggedright\arraybackslash}p{(\columnwidth - 6\tabcolsep) * \real{0.5647}}@{}}
\caption{Performance of FastText and GloVe Moral Foundations Lexicons on
Clifford Vignettes}\tabularnewline
\toprule\noalign{}
\begin{minipage}[b]{\linewidth}\raggedright
Lexicon
\end{minipage} & \begin{minipage}[b]{\linewidth}\raggedleft
F1
\end{minipage} & \begin{minipage}[b]{\linewidth}\raggedleft
Accuracy
\end{minipage} & \begin{minipage}[b]{\linewidth}\raggedright
Test
\end{minipage} \\
\midrule\noalign{}
\endfirsthead
\toprule\noalign{}
\begin{minipage}[b]{\linewidth}\raggedright
Lexicon
\end{minipage} & \begin{minipage}[b]{\linewidth}\raggedleft
F1
\end{minipage} & \begin{minipage}[b]{\linewidth}\raggedleft
Accuracy
\end{minipage} & \begin{minipage}[b]{\linewidth}\raggedright
Test
\end{minipage} \\
\midrule\noalign{}
\endhead
\bottomrule\noalign{}
\endlastfoot
MoralBERT & 0.384673 & 0.384673 & Moral Frame Identification (Clifford
Vignettes) \\
EMFD & 0.302467 & 0.302467 & Moral Frame Identification (Clifford
Vignettes) \\
FastText & 0.296474 & 0.296474 & Moral Frame Identification (Clifford
Vignettes) \\
GloVe & 0.145011 & 0.145011 & Moral Frame Identification (Clifford
Vignettes) \\
MFD & 0.315179 & 0.315179 & Moral Frame Identification (Clifford
Vignettes) \\
MoralStrength v1.1 & 0.139971 & 0.139971 & Moral Frame Identification
(Clifford Vignettes) \\
MoralStrength v1 & 0.107750 & 0.107750 & Moral Frame Identification
(Clifford Vignettes) \\
GPT 4o & 0.022222 & 0.022222 & Moral Frame Identification (Clifford
Vignettes) \\
\end{longtable}

\paragraph{Reddit Forum Posts}\label{reddit-forum-posts}

\begin{longtable}[]{@{}
  >{\raggedright\arraybackslash}p{(\columnwidth - 8\tabcolsep) * \real{0.0395}}
  >{\raggedright\arraybackslash}p{(\columnwidth - 8\tabcolsep) * \real{0.2500}}
  >{\raggedleft\arraybackslash}p{(\columnwidth - 8\tabcolsep) * \real{0.1184}}
  >{\raggedleft\arraybackslash}p{(\columnwidth - 8\tabcolsep) * \real{0.1184}}
  >{\raggedright\arraybackslash}p{(\columnwidth - 8\tabcolsep) * \real{0.4737}}@{}}
\caption{Performance of FastText and GloVe Moral Foundations Lexicons on
Reddit Moral Foundations Corpus (Average across 3
Forums)}\tabularnewline
\toprule\noalign{}
\begin{minipage}[b]{\linewidth}\raggedright
\end{minipage} & \begin{minipage}[b]{\linewidth}\raggedright
Lexicon
\end{minipage} & \begin{minipage}[b]{\linewidth}\raggedleft
F1
\end{minipage} & \begin{minipage}[b]{\linewidth}\raggedleft
Accuracy
\end{minipage} & \begin{minipage}[b]{\linewidth}\raggedright
Test
\end{minipage} \\
\midrule\noalign{}
\endfirsthead
\toprule\noalign{}
\begin{minipage}[b]{\linewidth}\raggedright
\end{minipage} & \begin{minipage}[b]{\linewidth}\raggedright
Lexicon
\end{minipage} & \begin{minipage}[b]{\linewidth}\raggedleft
F1
\end{minipage} & \begin{minipage}[b]{\linewidth}\raggedleft
Accuracy
\end{minipage} & \begin{minipage}[b]{\linewidth}\raggedright
Test
\end{minipage} \\
\midrule\noalign{}
\endhead
\bottomrule\noalign{}
\endlastfoot
4 & GPT 4o & 0.267154 & 0.267154 & Moral Frame Identification
(Reddit) \\
5 & MFD & 0.260299 & 0.260299 & Moral Frame Identification (Reddit) \\
6 & MoralBERT & 0.273573 & 0.273573 & Moral Frame Identification
(Reddit) \\
2 & FastText & 0.137601 & 0.137601 & Moral Frame Identification
(Reddit) \\
1 & EMFD & 0.095884 & 0.095884 & Moral Frame Identification (Reddit) \\
8 & MoralStrength v1.1 & 0.252075 & 0.252075 & Moral Frame
Identification (Reddit) \\
3 & GloVe & 0.083569 & 0.083569 & Moral Frame Identification (Reddit) \\
7 & MoralStrength v1 & 0.253761 & 0.253761 & Moral Frame Identification
(Reddit) \\
\end{longtable}

\paragraph{News Articles}\label{news-articles}

\begin{longtable}[]{@{}
  >{\raggedright\arraybackslash}p{(\columnwidth - 8\tabcolsep) * \real{0.0333}}
  >{\raggedright\arraybackslash}p{(\columnwidth - 8\tabcolsep) * \real{0.2111}}
  >{\raggedleft\arraybackslash}p{(\columnwidth - 8\tabcolsep) * \real{0.1000}}
  >{\raggedleft\arraybackslash}p{(\columnwidth - 8\tabcolsep) * \real{0.1000}}
  >{\raggedright\arraybackslash}p{(\columnwidth - 8\tabcolsep) * \real{0.5556}}@{}}
\caption{Performance of FastText and GloVe Moral Foundations Lexicons on
U.S. News Articles, Document-Level}\tabularnewline
\toprule\noalign{}
\begin{minipage}[b]{\linewidth}\raggedright
\end{minipage} & \begin{minipage}[b]{\linewidth}\raggedright
Lexicon
\end{minipage} & \begin{minipage}[b]{\linewidth}\raggedleft
F1
\end{minipage} & \begin{minipage}[b]{\linewidth}\raggedleft
Accuracy
\end{minipage} & \begin{minipage}[b]{\linewidth}\raggedright
Test
\end{minipage} \\
\midrule\noalign{}
\endfirsthead
\toprule\noalign{}
\begin{minipage}[b]{\linewidth}\raggedright
\end{minipage} & \begin{minipage}[b]{\linewidth}\raggedright
Lexicon
\end{minipage} & \begin{minipage}[b]{\linewidth}\raggedleft
F1
\end{minipage} & \begin{minipage}[b]{\linewidth}\raggedleft
Accuracy
\end{minipage} & \begin{minipage}[b]{\linewidth}\raggedright
Test
\end{minipage} \\
\midrule\noalign{}
\endhead
\bottomrule\noalign{}
\endlastfoot
9 & GloVe & 0.290844 & 0.290844 & Moral Frame Identification (News,
Document-Level) \\
10 & MoralBERT & 0.369597 & 0.369597 & Moral Frame Identification (News,
Document-Level) \\
11 & EMFD & 0.351883 & 0.351883 & Moral Frame Identification (News,
Document-Level) \\
12 & GPT 4o & 0.279925 & 0.279925 & Moral Frame Identification (News,
Document-Level) \\
13 & MFD & 0.281135 & 0.281135 & Moral Frame Identification (News,
Document-Level) \\
14 & FastText & 0.216691 & 0.216691 & Moral Frame Identification (News,
Document-Level) \\
15 & MoralStrength v1.1 & 0.238195 & 0.238195 & Moral Frame
Identification (News, Document-Level) \\
16 & MoralStrength v1 & 0.349469 & 0.349469 & Moral Frame Identification
(News, Document-Level) \\
\end{longtable}

\begin{longtable}[]{@{}
  >{\raggedright\arraybackslash}p{(\columnwidth - 8\tabcolsep) * \real{0.0349}}
  >{\raggedright\arraybackslash}p{(\columnwidth - 8\tabcolsep) * \real{0.2209}}
  >{\raggedleft\arraybackslash}p{(\columnwidth - 8\tabcolsep) * \real{0.1047}}
  >{\raggedleft\arraybackslash}p{(\columnwidth - 8\tabcolsep) * \real{0.1047}}
  >{\raggedright\arraybackslash}p{(\columnwidth - 8\tabcolsep) * \real{0.5349}}@{}}
\caption{Performance of FastText and GloVe Moral Foundations Lexicons on
U.S. News Articles, Line-Level}\tabularnewline
\toprule\noalign{}
\begin{minipage}[b]{\linewidth}\raggedright
\end{minipage} & \begin{minipage}[b]{\linewidth}\raggedright
Lexicon
\end{minipage} & \begin{minipage}[b]{\linewidth}\raggedleft
F1
\end{minipage} & \begin{minipage}[b]{\linewidth}\raggedleft
Accuracy
\end{minipage} & \begin{minipage}[b]{\linewidth}\raggedright
Test
\end{minipage} \\
\midrule\noalign{}
\endfirsthead
\toprule\noalign{}
\begin{minipage}[b]{\linewidth}\raggedright
\end{minipage} & \begin{minipage}[b]{\linewidth}\raggedright
Lexicon
\end{minipage} & \begin{minipage}[b]{\linewidth}\raggedleft
F1
\end{minipage} & \begin{minipage}[b]{\linewidth}\raggedleft
Accuracy
\end{minipage} & \begin{minipage}[b]{\linewidth}\raggedright
Test
\end{minipage} \\
\midrule\noalign{}
\endhead
\bottomrule\noalign{}
\endlastfoot
17 & MoralBERT & 0.396721 & 0.396721 & Moral Frame Identification (News,
Line-Level) \\
18 & EMFD & 0.286264 & 0.286264 & Moral Frame Identification (News,
Line-Level) \\
19 & GloVe & 0.285260 & 0.285260 & Moral Frame Identification (News,
Line-Level) \\
20 & FastText & 0.294216 & 0.294216 & Moral Frame Identification (News,
Line-Level) \\
21 & MFD & 0.278121 & 0.278121 & Moral Frame Identification (News,
Line-Level) \\
22 & MoralStrength v1.1 & 0.279130 & 0.279130 & Moral Frame
Identification (News, Line-Level) \\
23 & GPT 4o & 0.141467 & 0.141467 & Moral Frame Identification (News,
Line-Level) \\
24 & MoralStrength v1 & 0.232360 & 0.232360 & Moral Frame Identification
(News, Line-Level) \\
\end{longtable}

\(\qquad\) As in sentiment tests, the generic embedding dictionaries
perform roughly similarly, and at times better, than state-of-the-art
models such as MoralBERT. MoralBERT is also explicitly trained on the
Reddit corpus tested, which would be expected to further boost its
performance. The FastText and GloVe lexicons are consistently moderate
to high in accuracy, compared to the higher performance variance seen
from LLM-based labelling or MoralBERT in some tests. Scoring texts with
a lexicon method is also much reduced in terms of computing time,
whereas a complex model such as MoralBERT can take 10 seconds per entry
and 1-2 hours for a moderately sized, 3000-entry social media post
dataset.

\hfill\break

\subsubsection{The Distinct Positive Emotion
Lexicon}\label{the-distinct-positive-emotion-lexicon}

\hfill\break

\(\qquad\) Given that labelled datasets are limited in availability and
scope, an alternative test of how well the generic embedding lexicons
perform is to compare their similarity (in word feature contents) to
existing validated lexicons, or to compare the scores they produce when
applied to familiar texts, such as speeches from well-known political
figures.

\(\qquad\) In response, the final lexicon presented measures 7 distinct
positive emotions in U.S. presidents' inaugural speeches since Jimmy
Carter's (1977; the speeches are sourced from Quanteda). Reducing a list
of 10 positive emotions found in Zomeren (2021) to six, and adding
nostalgia as a politically relevant emotion (Bonansinga 2020), the
emotions captured are Awe (or Aesthetic Appreciation); Joy or Amusement
(in the sense of playful, high-energy positive emotion); Hope; Love;
Nostalgia; Pride; and Serenity or Contentment (low-energy, calm positive
emotion). For each positive emotion, 1500 words representing the
positive pole and 500 its opposite are retained. (Only positive pole
words are illustrated below).

\(\qquad\) Previous theoretical work on emotion has pointed to a lack of
attention to disambiguating positive emotion, in comparison to the
greater focus on distinguishing moods and sensations at the unpleasant
pole (see, for example, the typologies in Ortony (2022), p.48).
Literature on political communication indicates a need for greater
detail than many distinct-emotion lexicons (which are relatively rare;
NRC Emotion Lexicon is perhaps the most prominent or only readily
available one). Right-leaning populist figures, for example, are thought
to invoke nostalgia (Bonansinga 2020), aesthetic appreciation, hope,
appeals to religion and tradition, and hedonism (jokes, sarcasm)
(Albertazzi and Bonansinga 2024) as the more positive emotions and
framings they employ.

\(\qquad\) Per popular understandings of recent presidential campaigns,
we might therefore expect Barack Obama's first address to speak
especially strongly of `hope', and for Trump's following the 2016
election to elicit nostalgic feeling.\footnote{As with the other
  lexicons presented, please note that the words associated with a
  concept have this connotation based on their shared use, in close
  proximity to words of similar meaning or associations, in the training
  or `estimation' corpus. This meaning-attributing process is thus
  different from requiring human annotators to determine a word has
  particular connotations, as in the approach taken to construct the NRC
  Emotion Lexicon or Lexicoder. Rather, corpus frequency statistics are
  the basis of words being positioned more closely to one conceptual
  pole or another, or to neither, in large quantities of `natural'
  written language.}

\hfill\break

\paragraph{Distinct Positive Emotions in U.S. Presidents' Inaugural
Speeches}\label{distinct-positive-emotions-in-u.s.-presidents-inaugural-speeches}

\hfill\break

\(\qquad\) As expected, Trump's inaugural speech of early 2017 scores
the highest for the presence of nostalgia, as well as among the highest
for awe, hope, and pride. An especially high-energy (joyful, not
`serene') speech is Bill Clinton's of 1997, and the least `proud' are
delivered by Joe Biden in 2021 and by Barack Obama in 2009. Other
exceptionally low or high scoring speeches include George W. Bush's low
prevalence of `love' words in 2005, likely relating to ongoing wars. (An
illustration of scores by speech appears in the Appendix).

\(\qquad\) Excerpts follow of particularly low- or high-scoring
paragraphs for these emotions, within a possible -1 to 1 range. All are
scored and ranked by the FastText dictionary.

\begin{longtable}[]{@{}
  >{\raggedright\arraybackslash}p{(\columnwidth - 4\tabcolsep) * \real{0.2222}}
  >{\raggedright\arraybackslash}p{(\columnwidth - 4\tabcolsep) * \real{0.4583}}
  >{\raggedright\arraybackslash}p{(\columnwidth - 4\tabcolsep) * \real{0.1944}}@{}}
\caption{High-Scoring Positive Emotion Paragraphs, U.S. Inaugural
Addresses}\tabularnewline
\toprule\noalign{}
\begin{minipage}[b]{\linewidth}\raggedright
Emotion
\end{minipage} & \begin{minipage}[b]{\linewidth}\raggedright
Text
\end{minipage} & \begin{minipage}[b]{\linewidth}\raggedright
Valence
\end{minipage} \\
\midrule\noalign{}
\endfirsthead
\toprule\noalign{}
\begin{minipage}[b]{\linewidth}\raggedright
Emotion
\end{minipage} & \begin{minipage}[b]{\linewidth}\raggedright
Text
\end{minipage} & \begin{minipage}[b]{\linewidth}\raggedright
Valence
\end{minipage} \\
\midrule\noalign{}
\endhead
\bottomrule\noalign{}
\endlastfoot
Awe & It is the American story - a story of flawed and fallible people,
united across the generations by grand and enduring ideals. (Bush, 2001)
& 1 (Highest) \\
Hope & On this day, we gather because we have chosen hope over fear,
unity of purpose over conflict and discord. (Obama, 2009) & 0.76
(3rd) \\
Joy/Amusement & Let us learn together and laugh together and work
together and pray together, confident that in the end we will triumph
together in the right. (Carter, 1977) & 0.76 (4th) \\
Joy/Amusement & We will build new roads, and highways, and bridges, and
airports, and tunnels, and railways all across our wonderful nation.
(Trump, 2017) & 0.75 (6th) \\
Love & Sometimes in life we are called to do great things. But as a
saint of our times has said, every day we are called to do small things
with great love. The most important tasks of a democracy are done by
everyone. (Bush, 2001) & 1 \\
Nostalgia & {[}\ldots{]} But my thoughts have been turning the past few
days to those who would be watching at home to an older fellow who will
throw a salute by himself when the flag goes by, and the women who will
tell her sons the words of the battle hymns. I don't mean this to be
sentimental. I mean that on days like this, we remember that we are all
part of a continuum, inescapably connected by the ties that bind. (Bush,
1989) & 0.69 (2nd) \\
Nostalgia & The forgotten men and women of our country will be forgotten
no longer. (Trump, 2017) & 0.63 (3rd) \\
Pride & We can deliver racial justice. (Biden, 2021) & 1 \\
Serenity & America, at its best, is compassionate. In the quiet of
American conscience, we know that deep, persistent poverty is unworthy
of our nation's promise. (Bush, 2001) & 1 \\
\end{longtable}

\newpage

\begin{longtable}[]{@{}
  >{\raggedright\arraybackslash}p{(\columnwidth - 4\tabcolsep) * \real{0.3291}}
  >{\raggedright\arraybackslash}p{(\columnwidth - 4\tabcolsep) * \real{0.4177}}
  >{\raggedright\arraybackslash}p{(\columnwidth - 4\tabcolsep) * \real{0.2532}}@{}}
\caption{Low-Scoring Positive Emotion Paragraphs, U.S. Inaugural
Addresses}\tabularnewline
\toprule\noalign{}
\begin{minipage}[b]{\linewidth}\raggedright
Emotion
\end{minipage} & \begin{minipage}[b]{\linewidth}\raggedright
Text
\end{minipage} & \begin{minipage}[b]{\linewidth}\raggedright
Valence
\end{minipage} \\
\midrule\noalign{}
\endfirsthead
\toprule\noalign{}
\begin{minipage}[b]{\linewidth}\raggedright
Emotion
\end{minipage} & \begin{minipage}[b]{\linewidth}\raggedright
Text
\end{minipage} & \begin{minipage}[b]{\linewidth}\raggedright
Valence
\end{minipage} \\
\midrule\noalign{}
\endhead
\bottomrule\noalign{}
\endlastfoot
(Against) Awe & It is time to break the bad habit of expecting something
for nothing: from our government, or from each other. {[}\ldots{]}
Powerful people maneuver for position and worry endlessly about who is
in and who is out, who is up and who is down, forgetting those people
whose toil and sweat sends us here and paves our way. (Clinton, 1993) &
-0.48 (Lowest) \\
(Against) Joy/Amusement & My most solemn duty is to protect this Nation
and its people from further attacks and emerging threats. {[}\ldots{]}
(Bush, 2005) & -1 \\
(Against) Love & For as long as whole regions of the world simmer in
resentment and tyranny, prone to ideologies that feed hatred and excuse
murder, violence will gather and multiply in destructive power and cross
the most defended borders and raise a mortal threat. {[}\ldots{]} (Bush,
2005) & -0.8 (2nd lowest) \\
(Against) Nostalgia & From this day forward, it's going to be only
America first, America first. (Trump, 2017) & -1 \\
(Against) Pride & As we consider the road that unfolds before us, we
remember with humble gratitude those brave Americans who, at this very
hour, patrol far-off deserts and distant mountains. {[}\ldots{]} (Obama,
2009) & -1 \\
(Against) Serenity & Then, in turmoil and triumph, that promise exploded
onto the world stage to make this the American Century. (Clinton, 1997)
& -1 \\
\end{longtable}

\newpage

\subsubsection{Methodological
Considerations}\label{methodological-considerations}

\hfill\break

\paragraph{Composition of the Embedding Training
Corpus}\label{composition-of-the-embedding-training-corpus}

\hfill\break

\(\qquad\) FastText's Common Crawl training data differs from the GloVe
corpus inputs, in that it appears to be much more informal and
uncensored. The 6B-trained GloVe, based on Wikipedia and Gigaword,
contains more politics- and news-relevant words among its most used
terms, and fewer related to web activities. As shown below, slang and
obscenities are more frequently used in the FastText training corpus, as
are words relating to online shopping and HTML. The larger vocabulary
and the greater prevalence of social and emotional language in the
FastText corpus may therefore make its lexicons more suitable than GloVe
for most texts, although GloVe appears to perform well in some tests at
higher formality levels.\footnote{Versions of GloVe based on Common
  Crawl data have a similar vocabulary to the FastText vectors used,
  meaning that the `training' data rather than the estimation algorithm
  is likely responsible for most differences in the outputted lexicons
  and word representations.}

\begin{figure}
\centering
\includegraphics{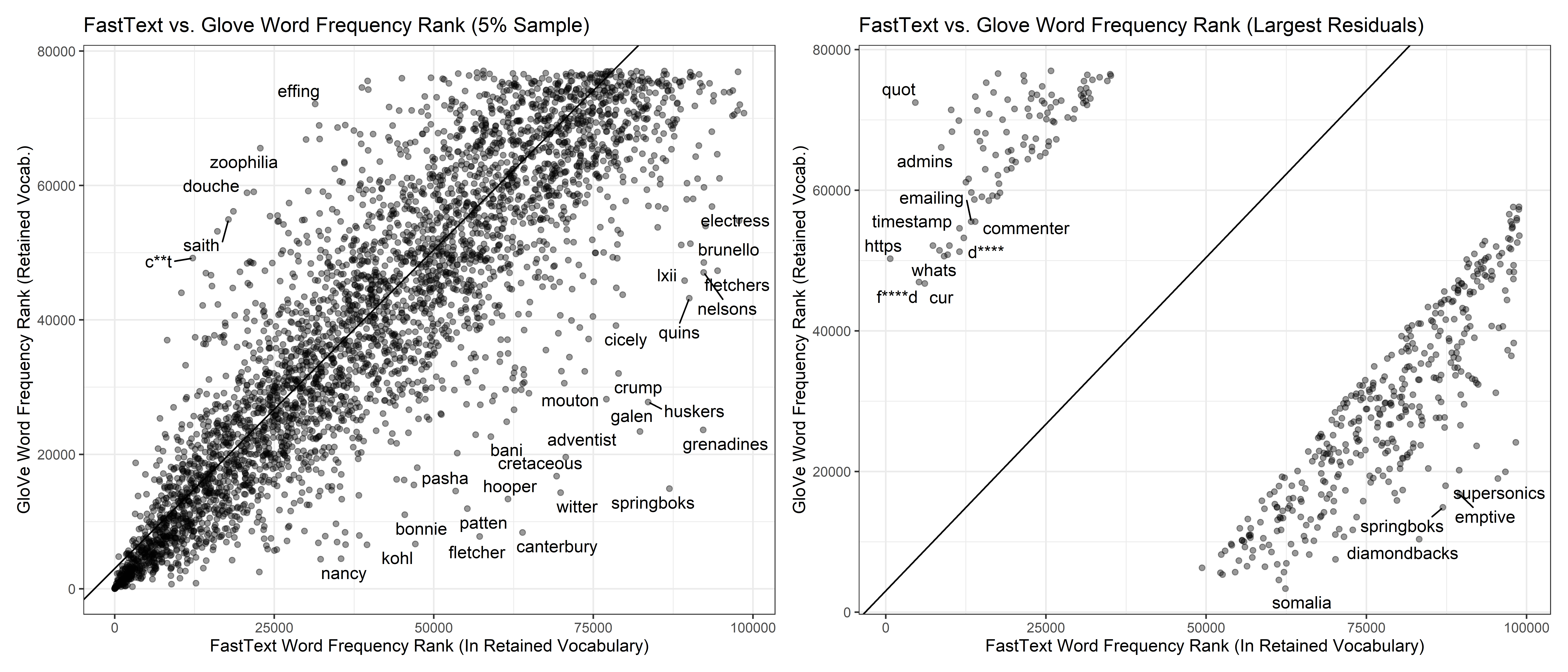}
\caption{Higher discrepancy word usage rates between the GloVe and
FastText training corpuses}
\end{figure}

\(\qquad\) The relevance of the training or estimation corpus is not in
word frequencies alone, though, but rather in how differences in inputs
may affect the conceptual associations of words. Garg et al. (2018) and
others have written about how large, human-written collections of input
texts can encapsulate gender and racial stereotypes that translate to
embedding representations' positioning. With the unfiltered full set of
GloVe 6B embeddings, for example, certain country names are returned as
close to the negative sentiment pole, indicating similarly fraught
associations. This concern is, to be certain, present in the use of
customized embeddings or LLMs as well, unbeknownst to the used. Manual
review of the (easily visible) lexicon words may be most suitable for
ensuring that no obviously inappropriate conceptual linkages are made.

\(\qquad\) Reassuringly, however, GloVe and FastText sentiment
associations with words do not vary greatly. After estimating each word
in the full vocabulary's sentiment valence, and regressing FastText
sentiment valence (-1 to 1 range) on GloVe's, no words have a residual
larger than 0.6 in size; and only 16 words' residuals are more extreme
than 0.5. Compared to the custom embeddings produced from the British
Hansard in Rheault et al. (2016), however, some gaps in sentiment
association (with either the FastText or GloVe valence) are present.
These may be due to changing word connotations over time (as the custom
Hansard embeddings start in 1909), or other particularities in word
usage in Parliament: `fine', `brand', and `founder' appear to be used in
their more negative-connoted verb forms, for example.\footnote{The
  implication is that running analyses on historic texts may not be well
  suited to the use of web-trained vector representations and custom
  embeddings may be more suitable, as in Schwartzberg and Spirling
  (n.d.)' application with early modern English texts to analyze
  Leveller thought.}

\begin{figure}
\centering
\includegraphics{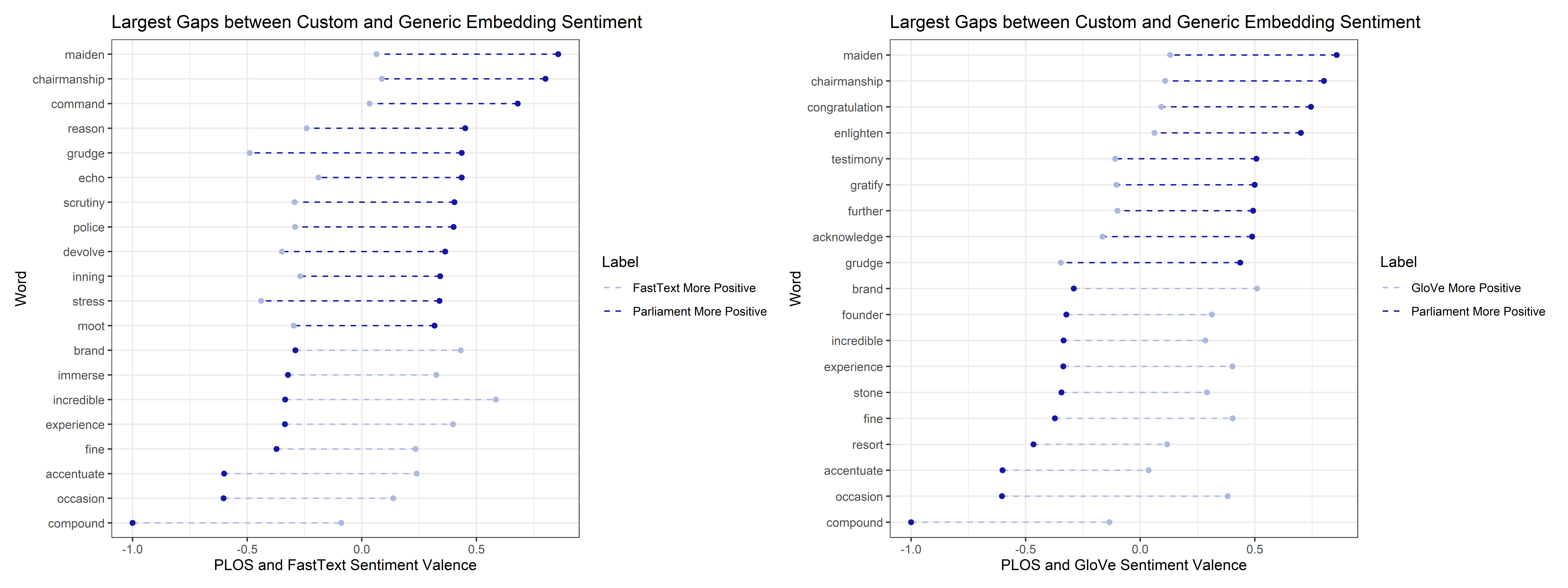}
\caption{Discrepancies in sentiment valence between generic and custom
embeddings}
\end{figure}

\(\qquad\) Likewise, similarities in both word composition and valence
are generally high between the FastText and GloVe sentiment lexicons and
crowdsourced alternatives Lexicoder and the NRC Emotion Lexicon. A full
40.5\% of returned words (excluding seeds, which match precisely) in the
GloVe lexicon are also found in the customized embedding lexicon shown
above, as do 32.5\% of the returned words for the FastText dictionary.
With Lexicoder, 22.9\% of the FastText dictionary and 25.3\% of the
GloVe lexicon overlap. Compared to the NRC Emotion Lexicon, FastText has
21.1\% overlap in words and GloVe 25.3\%. Agreement on the direction of
sentiment words is also high, at 96\% and 97\% of terms being placed in
the same `Positive' or `Negative' category among overlapping words in
FastText and GloVe respectively. Compared against Lexicoder's
categories, 90\% of FastText words and 87\% of GloVe's similarly align.
(A handful of terms with higher discrepancies in valence, as in the
figure above, appear for Lexicoder and the NRC lexicons in the
Appendix).

\hfill\break

\paragraph{Number of Seed Words
Needed}\label{number-of-seed-words-needed}

\hfill\break

\(\qquad\) One unknown quantity is how many seed words, at each pole,
are needed in order to populate a functional (relatively high accuracy)
and stable (consistent and similar in content) lexicon. To explore this
question, the moral frames lexicons for FastText and GloVe are produced
under a varying number, and type, of initial seed words. The final
versions presented above use a full list of 100 seeds at each pole; the
variants explored below have either 5 theoretical words at each pole,
core conceptual words drawn from Hoover et al. (2020), or random samples
of 5, 10, 25, 50, and 75 seed words at each pole from the `full list' of
100. (The longer lists also include MFD and EMFD lexicon terms). Each
sample's randomization `seed' is changed for each run, and between the
two embedding vocabularies, so the samples of 5, 10, 25, and so on do
not consist of the same words for FastText and GloVe.

\(\qquad\) Lexicon similarity by number of seeds is reported in the
Appendix; below, average performance and the standard deviation of
accuracy scores across six tests appear in decreasing order. Larger
numbers of seeds appear to produce a more reliable dictionary, in the
sense that performance variance is lower. At times, lexicons with fewer
seeds score higher, however. (Notably, as well, a lexicon produced from
5 carefully chosen `theoretical' words is as similar to the full
100-seed dictionary as one with 25 randomly chosen seeds, as shown in
the Appendix).

\begin{figure}
\centering
\includegraphics{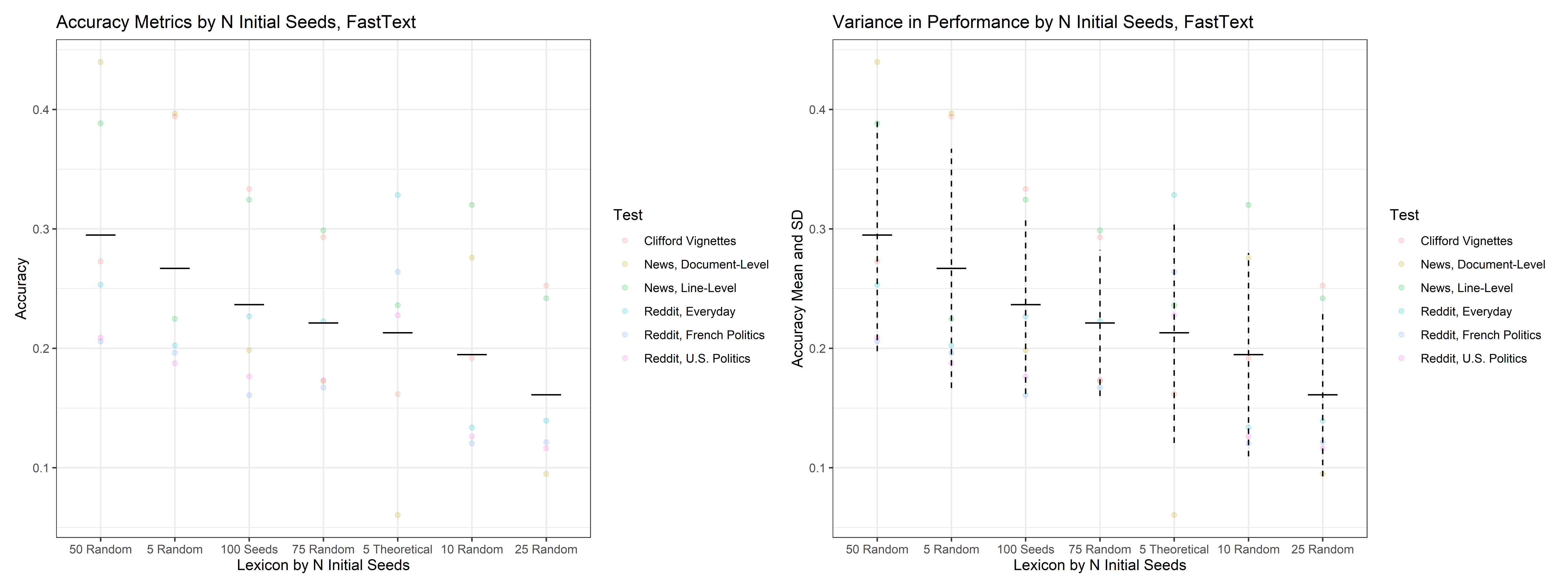}
\caption{Comparison of FastText lexicon performance by number of initial
seed words}
\end{figure}

\begin{figure}
\centering
\includegraphics{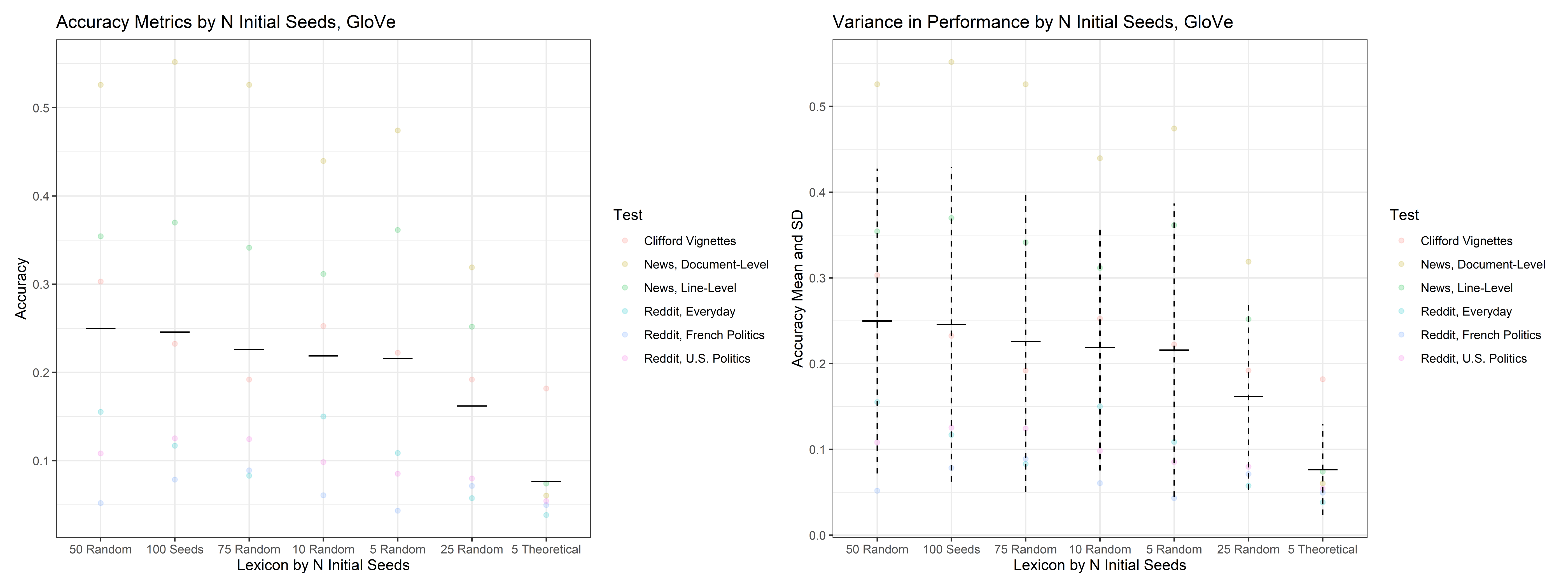}
\caption{Comparison of GloVe lexicon performance by number of initial
seed words}
\end{figure}

\(\qquad\) GloVe, in particular, which does not appear to have as large
of an emotional vocabulary as FastText, can produce weak performance
results with few seeds. GloVe performance is highest on more formal
texts, likely resulting from its more formal estimation data, coming
from an online encyclopedia and published news rather than social media
and general web sources, as FastText has.

\hfill\break

\subsection{Discussion}\label{discussion}

\paragraph{Benefits}\label{benefits}

\hfill\break

\(\qquad\) As noted throughout, generic embedding lexicons perform about
as well as, or better than, other current text scoring methods.
Crowdsourced lexicons and complex models perform extremely well in some
cases (Lexicoder for sentiment; GPT and MoralBERT for moral frames), but
fall short in others. Furthermore, needing extensive manual annotation
can limit the ability to create new text-measuring tools, as needed. The
time and cost inputs are substantial, compared to the similar
performance seen from the `semi-automatic', much lower-resource creation
of lexicons based on small lists of concept-defining words and available
generic embeddings. (Researcher checks for data quality from human
annotators are more time-consuming than the similar quality checks
needed on embedding-based lexicons, for example; training and hiring
annotators is another main cost).

\(\qquad\) Given transparency and other concerns with black-box models,
it is reassuring that generic embedding lexicons perform well. Users at
all levels of familiarity with text analysis tools can easily view what
words appear in the embedding lexicons, and could make modifications in
application, such as removing words that are extremely common (and do
not have any strong emotional connotations) from the lexicon or from the
corpus as needed. Results are transparently produced, in that only
single-word features influence the scores, and it is eminently possible
to check which words have the highest impact (for example, by subsetting
the term-document matrix to only words appearing in the lexicon used for
scoring, and summing column counts to identify which appear most
frequently throughout a corpus or document). Researchers are encouraged
to modify the lexicons (or their texts) accordingly, to avoid the
excessive impact of boilerplate language such as `Thank you', `Good
night', or in political documents phrases like `right honourable friend'
or `God bless America'.

\(\qquad\) The method is, additionally, highly flexible and open to the
development of new concept lexicons, other than those introduced here.
If high-performance and validated tools do not exist to label a
particular concept, or a researcher prefers a simpler approach,
producing a new lexicon is straightforward. Performance results across
the multiple lexicons tested (where competing lexicons and labelled data
do exist) suggest that the method returns conceptually relevant words,
and that word-scoring with an embedding based lexicon has competitively
strong performance to other leading models in applied tests.

\(\qquad\) Generic embedding lexicons do not rely excessively heavily on
a small group of researchers' impressions of word connotations, or
alternatively on large-scale, `crowdsourced' or trained coder annotation
processes. Rather, the semantic connections and associations implied by
word co-occurrences in large quantities of written texts inform the
vector representations used to populate the dictionaries. The high
quality of the lexicons, combined with their low-resource,
semi-automated creation, makes them a valuable tool for ongoing textual
research.

\hfill\break

\paragraph{Limitations and Future
Extensions}\label{limitations-and-future-extensions}

\hfill\break

\(\qquad\) Concerns over frequently occurring word features (in some
cases, effectively meaningless boilerplate) excessively influencing
scores exist with lexicon methods. However, given the high transparency
of the tools, frequently scored words can be easily identified, and
removed from the lexicon if deemed appropriate by researchers. Concerns
over the arbitrary dictionary size can be addressed either by valence
scoring (where the most conceptually extreme words will be most heavily
weighted), or, if producing a new lexicon, adjusting dictionary size
based on manual checks for the point at which word relevance to the
concept appears to become minimal.

\(\qquad\) The lexicons, as built (and as based on the format of the
pretrained word embeddings that are most readily available), base their
scoring on a unigram (single word) and `bag of words' (word order
irrelevant) approach. While recent works have found that unigram models
can perform similarly to those involving multi-word phrases (Araque,
Gatti, and Kalimeri 2020), it is theoretically possible to obtain
multi-word and/or sense-specific embeddings for the creation of similar
lexicons in the future. Alternatively, some researchers prefer to
compare the document embedding of a text's similarity to a lexicon
(Araque, Gatti, and Kalimeri 2020; Gennaro and Ash 2022). This method is
not the focus of the current paper, due to its slightly higher
complexity, but could be a valid option for those who do not prefer a
word-scoring approach. For the casual user, and even for more demanding
applications, unigram-based lexicon scores appear to perform relatively
well in providing appropriate measurements of texts.

\(\qquad\) Biases, whether social or statistical, can be present in any
of the methods applied; whether due to researcher prejudices in a
manually created lexicon, human annotator biases in a crowdsourced
alternative, or general societal biases in pretrained embeddings or LLMs
informed by large quantities of written texts. This phenomenon, however,
can most readily be identified and examined when the measuring
instrument is transparent in the features it uses to score a text toward
a certain conceptual direction. Rather than the `black box' nature of
many highly complex models, the lexicons presented are easily verifiable
in terms of their contents and the conceptual associations attributed to
each word.

\hfill\break

\subsection{Conclusions}\label{conclusions}

\hfill\break

\(\qquad\) Occupying a middle ground of complexity and performance,
generic embedding lexicons benefit from the information they encapsulate
through training (FastText) or estimation (GloVe) on statistical
regularities in (`large' scale) language usage across billions of tokens
of web and newswire text, while also enabling transparent measurement
methods that can be re-used across multiple corpuses and topical
domains, unlike existing customized models and opaque LLMs. They provide
users with a clear view of which word features push a document's score
in a certain direction, and can be lightly edited -- or created from
scratch -- based on the theoretical judgment, needs, and preferences of
the researcher.

\(\qquad\) In effect, the performance gains of increasingly complex,
often minimally transparent, models for text scoring are found to be
substantially small, or nonexistent. Throughout tests of performance --
based on popular and leading text methods and labelled data resources --
generic embedding lexicons perform well in correctly identifying the
emotional and other conceptual content of a text. For users looking to
apply the same measurement tool throughout multiple corpora, maximise
ease of use, obtain consistent and reliable measurements, and to have
full methodological transparency in how text scores are produced,
embedding-based lexicons provide many benefits with minimal downsides.

\mbox{} \vfill \textbf{Notes and Acknowledgments}

Thank you to the Centre for the Politics of Feelings (Royal Holloway and
School of Advanced Studies, University of London) for supporting this
research.

Comments are welcome at
\href{mailto:catherine.moez@rhul.ac.uk}{\nolinkurl{catherine.moez@rhul.ac.uk}}.

Released lexicons will be made available at
\url{https://github.com/catmoez/Generic-Embedding-Lexicons}.

\hfill\break

\newpage

\section{References}\label{references}

\phantomsection\label{refs}

\newpage

\section{Appendix}\label{appendix}

\subsection{Part 1: Dictionary
Contents}\label{part-1-dictionary-contents}

\subsubsection{Sentiment}\label{sentiment}

Extreme returned words in table format.

\begin{longtable}[]{@{}lllll@{}}
\caption{FastText, sentiment}\tabularnewline
\toprule\noalign{}
& word & valence & seed & sentiment \\
\midrule\noalign{}
\endfirsthead
\toprule\noalign{}
& word & valence & seed & sentiment \\
\midrule\noalign{}
\endhead
\bottomrule\noalign{}
\endlastfoot
1 & superb & 1 & 0 & Positive \\
2 & fantastic & 0.958 & 0 & Positive \\
3 & provide & 0.938 & 0 & Positive \\
4 & thoughtful & 0.907 & 0 & Positive \\
5 & dependable & 0.906 & 0 & Positive \\
6 & outstanding & 0.899 & 0 & Positive \\
7 & terrific & 0.887 & 0 & Positive \\
8 & exceptional & 0.882 & 0 & Positive \\
9 & versatile & 0.88 & 0 & Positive \\
10 & personable & 0.867 & 0 & Positive \\
11 & (\ldots) & & & \\
2790 & mishandling & -0.901 & 0 & Negative \\
2791 & plague & -0.903 & 0 & Negative \\
2792 & exacerbates & -0.911 & 0 & Negative \\
2793 & aggravates & -0.918 & 0 & Negative \\
2794 & crippling & -0.922 & 0 & Negative \\
2795 & inaction & -0.924 & 0 & Negative \\
2796 & intolerable & -0.931 & 0 & Negative \\
2797 & caused & -0.943 & 0 & Negative \\
2798 & mismanagement & -0.989 & 0 & Negative \\
2799 & blames & -0.993 & 0 & Negative \\
2800 & blamed & -1 & 0 & Negative \\
\end{longtable}

\newpage

\begin{longtable}[]{@{}lllll@{}}
\caption{GloVe, sentiment}\tabularnewline
\toprule\noalign{}
& word & valence & seed & sentiment \\
\midrule\noalign{}
\endfirsthead
\toprule\noalign{}
& word & valence & seed & sentiment \\
\midrule\noalign{}
\endhead
\bottomrule\noalign{}
\endlastfoot
1 & innovative & 1 & 0 & Positive \\
2 & elegant & 0.906 & 0 & Positive \\
3 & showcase & 0.906 & 0 & Positive \\
4 & versatile & 0.894 & 0 & Positive \\
5 & excellence & 0.888 & 0 & Positive \\
6 & unique & 0.877 & 0 & Positive \\
7 & enjoyed & 0.869 & 0 & Positive \\
8 & lively & 0.865 & 0 & Positive \\
9 & impressive & 0.855 & 0 & Positive \\
10 & appreciated & 0.845 & 0 & Positive \\
11 & (\ldots) & & & \\
2790 & exacerbating & -0.862 & 0 & Negative \\
2791 & blamed & -0.895 & 0 & Negative \\
2792 & causing & -0.9 & 0 & Negative \\
2793 & mishandling & -0.902 & 0 & Negative \\
2794 & blames & -0.903 & 0 & Negative \\
2795 & inaction & -0.922 & 0 & Negative \\
2796 & mismanagement & -0.926 & 0 & Negative \\
2797 & compounded & -0.931 & 0 & Negative \\
2798 & caused & -0.934 & 0 & Negative \\
2799 & blaming & -0.975 & 0 & Negative \\
2800 & exacerbated & -1 & 0 & Negative \\
\end{longtable}

\newpage

\subsubsection{Emotionality}\label{emotionality}

`Original' emotionality dictionary with 300+300 seed words. Please note
that the `original' dictionary leans slightly negative at the emotional
pole; alternatives are available.

\begin{longtable}[]{@{}lllll@{}}
\caption{FastText, emotionality (Original)}\tabularnewline
\toprule\noalign{}
& word & valence & seed & sentiment \\
\midrule\noalign{}
\endfirsthead
\toprule\noalign{}
& word & valence & seed & sentiment \\
\midrule\noalign{}
\endhead
\bottomrule\noalign{}
\endlastfoot
1 & softy & 1 & 0 & Affective \\
2 & miserable & 0.967 & 0 & Affective \\
3 & coward & 0.958 & 0 & Affective \\
4 & wimp & 0.958 & 0 & Affective \\
5 & heartbreak & 0.947 & 0 & Affective \\
6 & wuss & 0.944 & 0 & Affective \\
7 & misery & 0.925 & 0 & Affective \\
8 & sweetheart & 0.922 & 0 & Affective \\
9 & heartbreaker & 0.922 & 0 & Affective \\
10 & wicked & 0.92 & 0 & Affective \\
11 & (\ldots) & & & \\
2790 & verify & -0.842 & 0 & Cognitive \\
2791 & explicate & -0.842 & 0 & Cognitive \\
2792 & specifying & -0.85 & 0 & Cognitive \\
2793 & hypothesize & -0.865 & 0 & Cognitive \\
2794 & delineate & -0.875 & 0 & Cognitive \\
2795 & correspond & -0.893 & 0 & Cognitive \\
2796 & extrapolate & -0.896 & 0 & Cognitive \\
2797 & elucidate & -0.898 & 0 & Cognitive \\
2798 & indicate & -0.9 & 0 & Cognitive \\
2799 & ascertain & -0.998 & 0 & Cognitive \\
2800 & specify & -1 & 0 & Cognitive \\
\end{longtable}

\newpage

\begin{longtable}[]{@{}lllll@{}}
\caption{GloVe, emotionality (Original)}\tabularnewline
\toprule\noalign{}
& word & valence & seed & sentiment \\
\midrule\noalign{}
\endfirsthead
\toprule\noalign{}
& word & valence & seed & sentiment \\
\midrule\noalign{}
\endhead
\bottomrule\noalign{}
\endlastfoot
1 & endured & 1 & 0 & Affective \\
2 & suffering & 0.966 & 0 & Affective \\
3 & suffered & 0.941 & 0 & Affective \\
4 & misery & 0.938 & 0 & Affective \\
5 & beating & 0.919 & 0 & Affective \\
6 & weary & 0.919 & 0 & Affective \\
7 & crushing & 0.899 & 0 & Affective \\
8 & fierce & 0.89 & 0 & Affective \\
9 & amid & 0.888 & 0 & Affective \\
10 & miserable & 0.877 & 0 & Affective \\
11 & (\ldots) & & & \\
2790 & derivation & -0.916 & 0 & Cognitive \\
2791 & empirical & -0.918 & 0 & Cognitive \\
2792 & analogous & -0.923 & 0 & Cognitive \\
2793 & specified & -0.931 & 0 & Cognitive \\
2794 & elucidate & -0.934 & 0 & Cognitive \\
2795 & ascertain & -0.947 & 0 & Cognitive \\
2796 & specifying & -0.958 & 0 & Cognitive \\
2797 & correspond & -0.99 & 0 & Cognitive \\
2798 & parameters & -0.999 & 0 & Cognitive \\
2799 & postulated & -1 & 0 & Cognitive \\
2800 & hypotheses & -1 & 0 & Cognitive \\
\end{longtable}

\newpage

\paragraph{Fear and anger}\label{fear-and-anger}

\begin{longtable}[]{@{}lllll@{}}
\caption{FastText, fear and anger}\tabularnewline
\toprule\noalign{}
& word & valence & seed & sentiment \\
\midrule\noalign{}
\endfirsthead
\toprule\noalign{}
& word & valence & seed & sentiment \\
\midrule\noalign{}
\endhead
\bottomrule\noalign{}
\endlastfoot
1 & vile & 1 & 0 & Anger \\
2 & slander & 0.982 & 0 & Anger \\
3 & despicable & 0.964 & 0 & Anger \\
4 & vitriolic & 0.958 & 0 & Anger \\
5 & invective & 0.951 & 0 & Anger \\
6 & hatefulness & 0.93 & 0 & Anger \\
7 & insults & 0.925 & 0 & Anger \\
8 & hatefully & 0.898 & 0 & Anger \\
9 & invectives & 0.891 & 0 & Anger \\
10 & spitefulness & 0.889 & 0 & Anger \\
11 & (\ldots) & & & \\
2790 & panicking & -0.846 & 0 & Fear \\
2791 & chancy & -0.878 & 0 & Fear \\
2792 & nearing & -0.895 & 0 & Fear \\
2793 & future & -0.907 & 0 & Fear \\
2794 & dicey & -0.907 & 0 & Fear \\
2795 & stability & -0.925 & 0 & Fear \\
2796 & horizon & -0.925 & 0 & Fear \\
2797 & jitters & -0.949 & 0 & Fear \\
2798 & precariousness & -0.957 & 0 & Fear \\
2799 & looming & -0.978 & 0 & Fear \\
2800 & uncertainty & -1 & 0 & Fear \\
\end{longtable}

\newpage

\begin{longtable}[]{@{}lllll@{}}
\caption{GloVe, fear and anger}\tabularnewline
\toprule\noalign{}
& word & valence & seed & sentiment \\
\midrule\noalign{}
\endfirsthead
\toprule\noalign{}
& word & valence & seed & sentiment \\
\midrule\noalign{}
\endhead
\bottomrule\noalign{}
\endlastfoot
1 & vindictive & 1 & 0 & Anger \\
2 & slander & 0.989 & 0 & Anger \\
3 & slur & 0.975 & 0 & Anger \\
4 & invective & 0.962 & 0 & Anger \\
5 & sexist & 0.927 & 0 & Anger \\
6 & despicable & 0.926 & 0 & Anger \\
7 & homophobic & 0.92 & 0 & Anger \\
8 & animus & 0.92 & 0 & Anger \\
9 & insults & 0.917 & 0 & Anger \\
10 & racist & 0.915 & 0 & Anger \\
11 & (\ldots) & & & \\
2790 & prospect & -0.77 & 0 & Fear \\
2791 & recession & -0.785 & 0 & Fear \\
2792 & crisis & -0.788 & 0 & Fear \\
2793 & possibility & -0.79 & 0 & Fear \\
2794 & downturn & -0.8 & 0 & Fear \\
2795 & worries & -0.809 & 0 & Fear \\
2796 & scenario & -0.829 & 0 & Fear \\
2797 & worried & -0.86 & 0 & Fear \\
2798 & uncertainties & -0.952 & 0 & Fear \\
2799 & uncertainty & -0.964 & 0 & Fear \\
2800 & risks & -1 & 0 & Fear \\
\end{longtable}

\begin{table}
\caption{\label{tab:unnamed-chunk-23}FastText Moral Frames: Authority, Care, Fairness, Loyalty}

\centering
\begin{tabular}[t]{l|l|l|l}
\hline
word & valence & seed & sentiment\\
\hline
bestow & 1 & 0 & Authority\\
\hline
guidance & 0.978 & 0 & Authority\\
\hline
obligates & 0.974 & 0 & Authority\\
\hline
appoint & 0.965 & 0 & Authority\\
\hline
confer & 0.954 & 0 & Authority\\
\hline
requires & 0.935 & 0 & Authority\\
\hline
delegated & 0.925 & 0 & Authority\\
\hline
entrusted & 0.922 & 0 & Authority\\
\hline
(...) &  &  & \\
\hline
fomenting & -0.803 & 0 & Anti-Authority\\
\hline
revolts & -0.806 & 0 & Anti-Authority\\
\hline
seditious & -0.809 & 0 & Anti-Authority\\
\hline
violent & -0.826 & 0 & Anti-Authority\\
\hline
unrest & -0.839 & 0 & Anti-Authority\\
\hline
insurrectionary & -0.86 & 0 & Anti-Authority\\
\hline
insurrections & -0.874 & 0 & Anti-Authority\\
\hline
revolt & -0.879 & 0 & Anti-Authority\\
\hline
rioting & -1 & 0 & Anti-Authority\\
\hline
\end{tabular}
\centering
\begin{tabular}[t]{l|l|l|l}
\hline
word & valence & seed & sentiment\\
\hline
supportive & 1 & 0 & Care\\
\hline
appreciated & 0.994 & 0 & Care\\
\hline
wonderful & 0.992 & 0 & Care\\
\hline
appreciation & 0.991 & 0 & Care\\
\hline
grateful & 0.979 & 0 & Care\\
\hline
thank & 0.951 & 0 & Care\\
\hline
thoughtfulness & 0.946 & 0 & Care\\
\hline
gratitude & 0.94 & 0 & Care\\
\hline
(...) &  &  & \\
\hline
inflicts & -0.909 & 0 & Anti-Care\\
\hline
terrorisation & -0.911 & 0 & Anti-Care\\
\hline
brutalising & -0.913 & 0 & Anti-Care\\
\hline
violences & -0.924 & 0 & Anti-Care\\
\hline
retaliations & -0.932 & 0 & Anti-Care\\
\hline
massacres & -0.942 & 0 & Anti-Care\\
\hline
brutalization & -0.959 & 0 & Anti-Care\\
\hline
terrorization & -0.993 & 0 & Anti-Care\\
\hline
harassments & -1 & 0 & Anti-Care\\
\hline
\end{tabular}
\centering
\begin{tabular}[t]{l|l|l|l}
\hline
word & valence & seed & sentiment\\
\hline
commitment & 1 & 0 & Fairness\\
\hline
adequate & 0.981 & 0 & Fairness\\
\hline
comprehensiveness & 0.968 & 0 & Fairness\\
\hline
satisfactory & 0.947 & 0 & Fairness\\
\hline
appropriate & 0.945 & 0 & Fairness\\
\hline
dedication & 0.938 & 0 & Fairness\\
\hline
requirement & 0.922 & 0 & Fairness\\
\hline
sufficient & 0.921 & 0 & Fairness\\
\hline
(...) &  &  & \\
\hline
deceitful & -0.94 & 0 & Anti-Fairness\\
\hline
conmen & -0.941 & 0 & Anti-Fairness\\
\hline
swindling & -0.942 & 0 & Anti-Fairness\\
\hline
thieving & -0.953 & 0 & Anti-Fairness\\
\hline
sleazes & -0.987 & 0 & Anti-Fairness\\
\hline
conman & -0.989 & 0 & Anti-Fairness\\
\hline
swindlers & -0.992 & 0 & Anti-Fairness\\
\hline
duped & -0.997 & 0 & Anti-Fairness\\
\hline
swindler & -1 & 0 & Anti-Fairness\\
\hline
\end{tabular}
\centering
\begin{tabular}[t]{l|l|l|l}
\hline
word & valence & seed & sentiment\\
\hline
brotherhood & 1 & 0 & Loyalty\\
\hline
togetherness & 0.96 & 0 & Loyalty\\
\hline
alumni & 0.945 & 0 & Loyalty\\
\hline
kinships & 0.936 & 0 & Loyalty\\
\hline
sisterhoods & 0.923 & 0 & Loyalty\\
\hline
sisterhood & 0.92 & 0 & Loyalty\\
\hline
alumnae & 0.916 & 0 & Loyalty\\
\hline
connectedness & 0.896 & 0 & Loyalty\\
\hline
(...) &  &  & \\
\hline
blamed & -0.872 & 0 & Anti-Loyalty\\
\hline
damaging & -0.872 & 0 & Anti-Loyalty\\
\hline
vindictive & -0.873 & 0 & Anti-Loyalty\\
\hline
blatantly & -0.889 & 0 & Anti-Loyalty\\
\hline
accusation & -0.891 & 0 & Anti-Loyalty\\
\hline
underhanded & -0.907 & 0 & Anti-Loyalty\\
\hline
untruthful & -0.907 & 0 & Anti-Loyalty\\
\hline
unethical & -0.917 & 0 & Anti-Loyalty\\
\hline
dishonest & -1 & 0 & Anti-Loyalty\\
\hline
\end{tabular}
\end{table}

\begin{table}
\caption{\label{tab:unnamed-chunk-23}FastText Moral Frames: Sanctity}

\centering
\begin{tabular}[t]{l|l|l|l}
\hline
word & valence & seed & sentiment\\
\hline
prayerfulness & 1 & 0 & Sanctity\\
\hline
prayerful & 0.938 & 0 & Sanctity\\
\hline
devotedness & 0.889 & 0 & Sanctity\\
\hline
faithfulness & 0.888 & 0 & Sanctity\\
\hline
devotion & 0.885 & 0 & Sanctity\\
\hline
scriptural & 0.879 & 0 & Sanctity\\
\hline
apostolic & 0.862 & 0 & Sanctity\\
\hline
intercession & 0.858 & 0 & Sanctity\\
\hline
(...) &  &  & \\
\hline
malodorous & -0.849 & 0 & Anti-Sanctity\\
\hline
vomit & -0.854 & 0 & Anti-Sanctity\\
\hline
festering & -0.86 & 0 & Anti-Sanctity\\
\hline
stench & -0.861 & 0 & Anti-Sanctity\\
\hline
fetid & -0.871 & 0 & Anti-Sanctity\\
\hline
smelly & -0.877 & 0 & Anti-Sanctity\\
\hline
rotting & -0.89 & 0 & Anti-Sanctity\\
\hline
infested & -0.949 & 0 & Anti-Sanctity\\
\hline
nasty & -1 & 0 & Anti-Sanctity\\
\hline
\end{tabular}
\end{table}

\begin{table}
\caption{\label{tab:unnamed-chunk-24}GloVe Moral Frames: Authority, Care, Fairness, Loyalty}

\centering
\begin{tabular}[t]{l|l|l|l}
\hline
word & valence & seed & sentiment\\
\hline
receive & 1 & 0 & Authority\\
\hline
required & 0.969 & 0 & Authority\\
\hline
must & 0.961 & 0 & Authority\\
\hline
give & 0.95 & 0 & Authority\\
\hline
chosen & 0.941 & 0 & Authority\\
\hline
appointed & 0.911 & 0 & Authority\\
\hline
given & 0.91 & 0 & Authority\\
\hline
appropriate & 0.903 & 0 & Authority\\
\hline
(...) &  &  & \\
\hline
rampages & -0.901 & 0 & Anti-Authority\\
\hline
presaged & -0.902 & 0 & Anti-Authority\\
\hline
degenerated & -0.902 & 0 & Anti-Authority\\
\hline
agitators & -0.906 & 0 & Anti-Authority\\
\hline
instigating & -0.911 & 0 & Anti-Authority\\
\hline
radicalism & -0.957 & 0 & Anti-Authority\\
\hline
rioting & -0.991 & 0 & Anti-Authority\\
\hline
insurrections & -0.991 & 0 & Anti-Authority\\
\hline
fomented & -1 & 0 & Anti-Authority\\
\hline
\end{tabular}
\centering
\begin{tabular}[t]{l|l|l|l}
\hline
word & valence & seed & sentiment\\
\hline
gift & 1 & 0 & Care\\
\hline
good & 0.992 & 0 & Care\\
\hline
thanks & 0.957 & 0 & Care\\
\hline
wonderful & 0.949 & 0 & Care\\
\hline
advice & 0.921 & 0 & Care\\
\hline
provide & 0.876 & 0 & Care\\
\hline
thank & 0.868 & 0 & Care\\
\hline
needs & 0.868 & 0 & Care\\
\hline
(...) &  &  & \\
\hline
grievous & -0.924 & 0 & Anti-Care\\
\hline
beatings & -0.93 & 0 & Anti-Care\\
\hline
rapes & -0.938 & 0 & Anti-Care\\
\hline
depredations & -0.947 & 0 & Anti-Care\\
\hline
inflicting & -0.95 & 0 & Anti-Care\\
\hline
mistreatment & -0.955 & 0 & Anti-Care\\
\hline
tortures & -0.956 & 0 & Anti-Care\\
\hline
injure & -0.958 & 0 & Anti-Care\\
\hline
pillage & -1 & 0 & Anti-Care\\
\hline
\end{tabular}
\centering
\begin{tabular}[t]{l|l|l|l}
\hline
word & valence & seed & sentiment\\
\hline
commitment & 1 & 0 & Fairness\\
\hline
necessary & 0.977 & 0 & Fairness\\
\hline
needed & 0.932 & 0 & Fairness\\
\hline
appropriate & 0.929 & 0 & Fairness\\
\hline
should & 0.929 & 0 & Fairness\\
\hline
maintain & 0.928 & 0 & Fairness\\
\hline
agreed & 0.928 & 0 & Fairness\\
\hline
expected & 0.924 & 0 & Fairness\\
\hline
(...) &  &  & \\
\hline
swindling & -0.884 & 0 & Anti-Fairness\\
\hline
tricking & -0.888 & 0 & Anti-Fairness\\
\hline
shamelessly & -0.892 & 0 & Anti-Fairness\\
\hline
dupes & -0.899 & 0 & Anti-Fairness\\
\hline
cahoots & -0.924 & 0 & Anti-Fairness\\
\hline
scheming & -0.932 & 0 & Anti-Fairness\\
\hline
scammed & -0.942 & 0 & Anti-Fairness\\
\hline
charlatan & -0.968 & 0 & Anti-Fairness\\
\hline
conned & -1 & 0 & Anti-Fairness\\
\hline
\end{tabular}
\centering
\begin{tabular}[t]{l|l|l|l}
\hline
word & valence & seed & sentiment\\
\hline
comprising & 1 & 0 & Loyalty\\
\hline
founded & 0.975 & 0 & Loyalty\\
\hline
alumni & 0.923 & 0 & Loyalty\\
\hline
formed & 0.91 & 0 & Loyalty\\
\hline
consisting & 0.887 & 0 & Loyalty\\
\hline
dedicated & 0.876 & 0 & Loyalty\\
\hline
established & 0.863 & 0 & Loyalty\\
\hline
partnership & 0.861 & 0 & Loyalty\\
\hline
(...) &  &  & \\
\hline
blatantly & -0.861 & 0 & Anti-Loyalty\\
\hline
unfounded & -0.863 & 0 & Anti-Loyalty\\
\hline
denounced & -0.873 & 0 & Anti-Loyalty\\
\hline
dismisses & -0.879 & 0 & Anti-Loyalty\\
\hline
harshly & -0.892 & 0 & Anti-Loyalty\\
\hline
deliberately & -0.898 & 0 & Anti-Loyalty\\
\hline
blackmail & -0.919 & 0 & Anti-Loyalty\\
\hline
accusations & -0.925 & 0 & Anti-Loyalty\\
\hline
accusation & -1 & 0 & Anti-Loyalty\\
\hline
\end{tabular}
\end{table}

\begin{table}
\caption{\label{tab:unnamed-chunk-24}GloVe Moral Frames: Sanctity}

\centering
\begin{tabular}[t]{l|l|l|l}
\hline
word & valence & seed & sentiment\\
\hline
church & 1 & 0 & Sanctity\\
\hline
tradition & 0.965 & 0 & Sanctity\\
\hline
dedication & 0.941 & 0 & Sanctity\\
\hline
catholic & 0.923 & 0 & Sanctity\\
\hline
dedicated & 0.916 & 0 & Sanctity\\
\hline
honor & 0.904 & 0 & Sanctity\\
\hline
congregation & 0.901 & 0 & Sanctity\\
\hline
devotion & 0.888 & 0 & Sanctity\\
\hline
(...) &  &  & \\
\hline
lice & -0.865 & 0 & Anti-Sanctity\\
\hline
megacolon & -0.87 & 0 & Anti-Sanctity\\
\hline
toxins & -0.873 & 0 & Anti-Sanctity\\
\hline
feces & -0.878 & 0 & Anti-Sanctity\\
\hline
bedbugs & -0.884 & 0 & Anti-Sanctity\\
\hline
infested & -0.886 & 0 & Anti-Sanctity\\
\hline
smelly & -0.9 & 0 & Anti-Sanctity\\
\hline
unsightly & -0.92 & 0 & Anti-Sanctity\\
\hline
infestation & -1 & 0 & Anti-Sanctity\\
\hline
\end{tabular}
\end{table}

\begin{table}
\caption{\label{tab:unnamed-chunk-25}FastText Distinct Positive Emotion Lexicon}

\centering
\begin{tabular}[t]{l|r|r|l}
\hline
word & valence & seed & sentiment\\
\hline
thrilled & 1.000 & 0 & Joy/Amusement\\
\hline
excited & 0.949 & 0 & Joy/Amusement\\
\hline
awesome & 0.883 & 0 & Joy/Amusement\\
\hline
giggle & 0.879 & 0 & Joy/Amusement\\
\hline
amazed & 0.874 & 0 & Joy/Amusement\\
\hline
adorable & 0.873 & 0 & Joy/Amusement\\
\hline
cute & 0.870 & 0 & Joy/Amusement\\
\hline
funny & 0.842 & 0 & Joy/Amusement\\
\hline
amazing & 0.803 & 0 & Joy/Amusement\\
\hline
happy & 0.795 & 0 & Joy/Amusement\\
\hline
\end{tabular}
\centering
\begin{tabular}[t]{l|r|r|l}
\hline
word & valence & seed & sentiment\\
\hline
awestruck & 1.000 & 0 & Awe\\
\hline
magnificence & 0.998 & 0 & Awe\\
\hline
majesty & 0.994 & 0 & Awe\\
\hline
marveled & 0.967 & 0 & Awe\\
\hline
admiration & 0.958 & 0 & Awe\\
\hline
marvel & 0.954 & 0 & Awe\\
\hline
marvelled & 0.951 & 0 & Awe\\
\hline
reverence & 0.922 & 0 & Awe\\
\hline
generosity & 0.906 & 0 & Awe\\
\hline
grandness & 0.905 & 0 & Awe\\
\hline
\end{tabular}
\centering
\begin{tabular}[t]{l|r|r|l}
\hline
word & valence & seed & sentiment\\
\hline
hopefully & 1.000 & 0 & Hope\\
\hline
envision & 0.989 & 0 & Hope\\
\hline
hoping & 0.981 & 0 & Hope\\
\hline
someday & 0.980 & 0 & Hope\\
\hline
intend & 0.904 & 0 & Hope\\
\hline
wait & 0.901 & 0 & Hope\\
\hline
hoped & 0.892 & 0 & Hope\\
\hline
anticipate & 0.873 & 0 & Hope\\
\hline
able & 0.872 & 0 & Hope\\
\hline
continue & 0.859 & 0 & Hope\\
\hline
\end{tabular}
\centering
\begin{tabular}[t]{l|r|r|l}
\hline
word & valence & seed & sentiment\\
\hline
sharing & 1.000 & 0 & Love\\
\hline
wonderful & 0.973 & 0 & Love\\
\hline
couples & 0.936 & 0 & Love\\
\hline
providing & 0.914 & 0 & Love\\
\hline
soulmate & 0.911 & 0 & Love\\
\hline
friendship & 0.904 & 0 & Love\\
\hline
keepsake & 0.902 & 0 & Love\\
\hline
bonding & 0.897 & 0 & Love\\
\hline
mentoring & 0.897 & 0 & Love\\
\hline
unforgettable & 0.891 & 0 & Love\\
\hline
\end{tabular}
\centering
\begin{tabular}[t]{l|r|r|l}
\hline
word & valence & seed & sentiment\\
\hline
boasts & 1.000 & 0 & Pride\\
\hline
deliver & 0.944 & 0 & Pride\\
\hline
ensure & 0.931 & 0 & Pride\\
\hline
ensured & 0.913 & 0 & Pride\\
\hline
unrivalled & 0.900 & 0 & Pride\\
\hline
ensuring & 0.860 & 0 & Pride\\
\hline
proudly & 0.837 & 0 & Pride\\
\hline
boasting & 0.835 & 0 & Pride\\
\hline
claimed & 0.831 & 0 & Pride\\
\hline
unparalleled & 0.825 & 0 & Pride\\
\hline
\end{tabular}
\centering
\begin{tabular}[t]{l|r|r|l}
\hline
word & valence & seed & sentiment\\
\hline
memories & 1.000 & 0 & Nostalgia\\
\hline
nostalgically & 0.949 & 0 & Nostalgia\\
\hline
remembrances & 0.933 & 0 & Nostalgia\\
\hline
reminisces & 0.930 & 0 & Nostalgia\\
\hline
reliving & 0.925 & 0 & Nostalgia\\
\hline
relive & 0.887 & 0 & Nostalgia\\
\hline
reminisced & 0.861 & 0 & Nostalgia\\
\hline
longing & 0.831 & 0 & Nostalgia\\
\hline
fondly & 0.814 & 0 & Nostalgia\\
\hline
yearning & 0.794 & 0 & Nostalgia\\
\hline
\end{tabular}
\centering
\begin{tabular}[t]{l|r|r|l}
\hline
word & valence & seed & sentiment\\
\hline
contentedness & 1.000 & 0 & Serenity\\
\hline
blissful & 0.995 & 0 & Serenity\\
\hline
peacefulness & 0.983 & 0 & Serenity\\
\hline
tranquility & 0.937 & 0 & Serenity\\
\hline
happiness & 0.902 & 0 & Serenity\\
\hline
blissfulness & 0.879 & 0 & Serenity\\
\hline
tranquillity & 0.861 & 0 & Serenity\\
\hline
restfulness & 0.855 & 0 & Serenity\\
\hline
restful & 0.822 & 0 & Serenity\\
\hline
nirvana & 0.817 & 0 & Serenity\\
\hline
\end{tabular}
\end{table}

\begin{table}
\caption{\label{tab:unnamed-chunk-26}GloVe Distinct Positive Emotion Dictionary}

\centering
\begin{tabular}[t]{l|r|r|l}
\hline
word & valence & seed & sentiment\\
\hline
liked & 1.000 & 0 & Joy/Amusement\\
\hline
laughing & 0.973 & 0 & Joy/Amusement\\
\hline
laughed & 0.934 & 0 & Joy/Amusement\\
\hline
funny & 0.911 & 0 & Joy/Amusement\\
\hline
happy & 0.910 & 0 & Joy/Amusement\\
\hline
excited & 0.890 & 0 & Joy/Amusement\\
\hline
everybody & 0.856 & 0 & Joy/Amusement\\
\hline
kids & 0.851 & 0 & Joy/Amusement\\
\hline
wonderful & 0.846 & 0 & Joy/Amusement\\
\hline
laughs & 0.844 & 0 & Joy/Amusement\\
\hline
\end{tabular}
\centering
\begin{tabular}[t]{l|r|r|l}
\hline
word & valence & seed & sentiment\\
\hline
statue & 1.000 & 0 & Awe\\
\hline
visited & 0.933 & 0 & Awe\\
\hline
greatest & 0.921 & 0 & Awe\\
\hline
great & 0.921 & 0 & Awe\\
\hline
admired & 0.881 & 0 & Awe\\
\hline
famed & 0.877 & 0 & Awe\\
\hline
renowned & 0.874 & 0 & Awe\\
\hline
park & 0.868 & 0 & Awe\\
\hline
heritage & 0.860 & 0 & Awe\\
\hline
celebrated & 0.830 & 0 & Awe\\
\hline
\end{tabular}
\centering
\begin{tabular}[t]{l|r|r|l}
\hline
word & valence & seed & sentiment\\
\hline
hoping & 1.000 & 0 & Hope\\
\hline
hoped & 0.919 & 0 & Hope\\
\hline
hopefully & 0.915 & 0 & Hope\\
\hline
expects & 0.908 & 0 & Hope\\
\hline
tomorrow & 0.860 & 0 & Hope\\
\hline
ready & 0.851 & 0 & Hope\\
\hline
hopes & 0.844 & 0 & Hope\\
\hline
chance & 0.830 & 0 & Hope\\
\hline
able & 0.830 & 0 & Hope\\
\hline
expected & 0.813 & 0 & Hope\\
\hline
\end{tabular}
\centering
\begin{tabular}[t]{l|r|r|l}
\hline
word & valence & seed & sentiment\\
\hline
friend & 1.000 & 0 & Love\\
\hline
couple & 0.997 & 0 & Love\\
\hline
wonderful & 0.955 & 0 & Love\\
\hline
lovely & 0.941 & 0 & Love\\
\hline
friends & 0.935 & 0 & Love\\
\hline
companion & 0.929 & 0 & Love\\
\hline
grace & 0.929 & 0 & Love\\
\hline
sister & 0.919 & 0 & Love\\
\hline
daughter & 0.906 & 0 & Love\\
\hline
wedding & 0.894 & 0 & Love\\
\hline
\end{tabular}
\centering
\begin{tabular}[t]{l|r|r|l}
\hline
word & valence & seed & sentiment\\
\hline
boasts & 1.000 & 0 & Pride\\
\hline
impressive & 0.936 & 0 & Pride\\
\hline
hopes & 0.906 & 0 & Pride\\
\hline
enjoyed & 0.877 & 0 & Pride\\
\hline
success & 0.872 & 0 & Pride\\
\hline
partnership & 0.856 & 0 & Pride\\
\hline
achieved & 0.852 & 0 & Pride\\
\hline
hoped & 0.850 & 0 & Pride\\
\hline
excellent & 0.849 & 0 & Pride\\
\hline
opportunity & 0.845 & 0 & Pride\\
\hline
\end{tabular}
\centering
\begin{tabular}[t]{l|r|r|l}
\hline
word & valence & seed & sentiment\\
\hline
longing & 1.000 & 0 & Nostalgia\\
\hline
relive & 0.983 & 0 & Nostalgia\\
\hline
bygone & 0.982 & 0 & Nostalgia\\
\hline
remembrances & 0.979 & 0 & Nostalgia\\
\hline
reverie & 0.972 & 0 & Nostalgia\\
\hline
revisits & 0.965 & 0 & Nostalgia\\
\hline
mawkish & 0.964 & 0 & Nostalgia\\
\hline
reminisces & 0.950 & 0 & Nostalgia\\
\hline
nostalgically & 0.944 & 0 & Nostalgia\\
\hline
pined & 0.924 & 0 & Nostalgia\\
\hline
\end{tabular}
\centering
\begin{tabular}[t]{l|r|r|l}
\hline
word & valence & seed & sentiment\\
\hline
solitude & 1.000 & 0 & Serenity\\
\hline
happiness & 0.999 & 0 & Serenity\\
\hline
secluded & 0.915 & 0 & Serenity\\
\hline
blissful & 0.896 & 0 & Serenity\\
\hline
everlasting & 0.889 & 0 & Serenity\\
\hline
peacefulness & 0.853 & 0 & Serenity\\
\hline
idyllic & 0.849 & 0 & Serenity\\
\hline
immortality & 0.839 & 0 & Serenity\\
\hline
blessedness & 0.837 & 0 & Serenity\\
\hline
goodness & 0.836 & 0 & Serenity\\
\hline
\end{tabular}
\end{table}

\hfill\break

\hfill\break

\hfill\break

\hfill\break

\hfill\break

\hfill\break

\hfill\break

\hfill\break

\newpage

\begin{figure}
\centering
\includegraphics{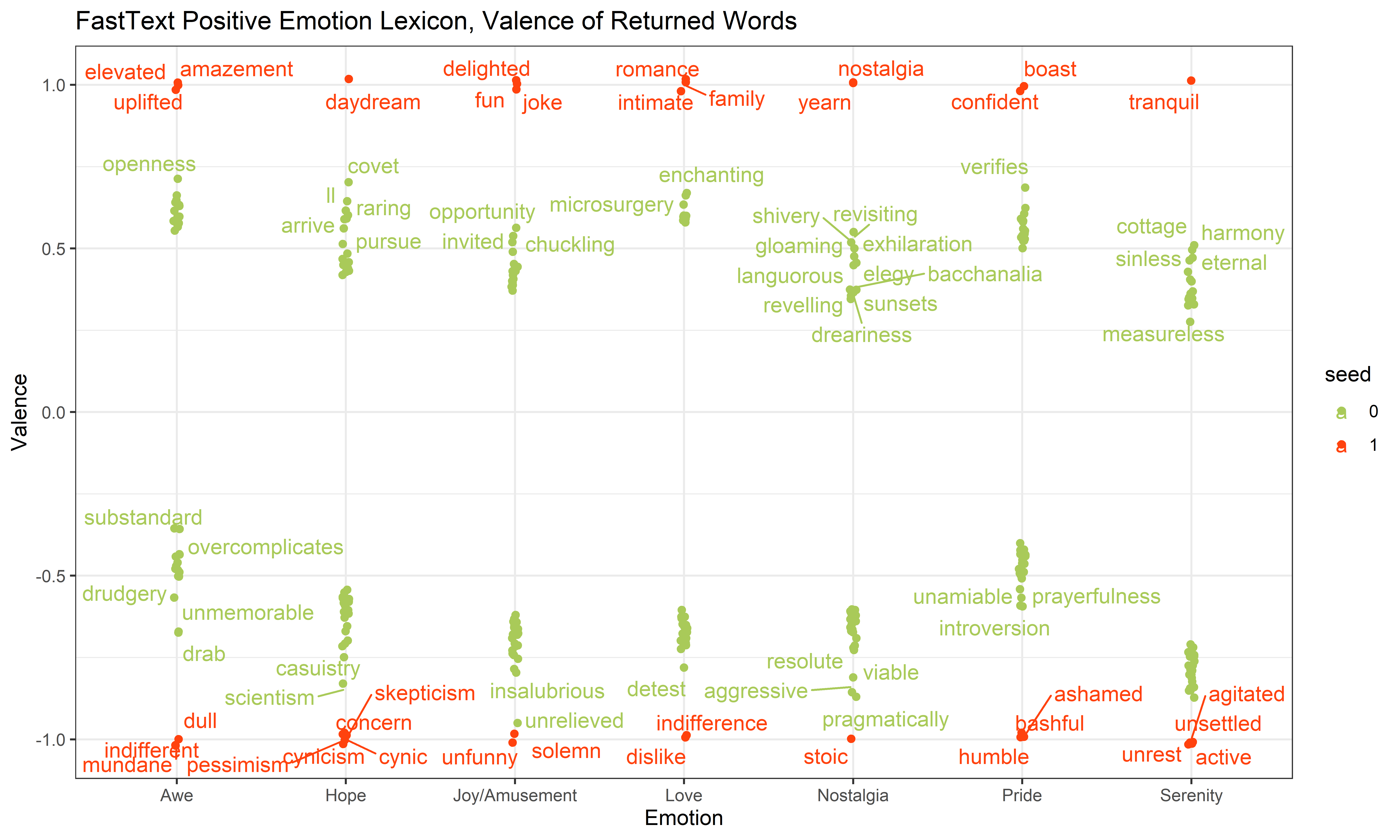}
\caption{Distinct positive emotion dictionary, FastText}
\end{figure}

\begin{figure}
\centering
\includegraphics{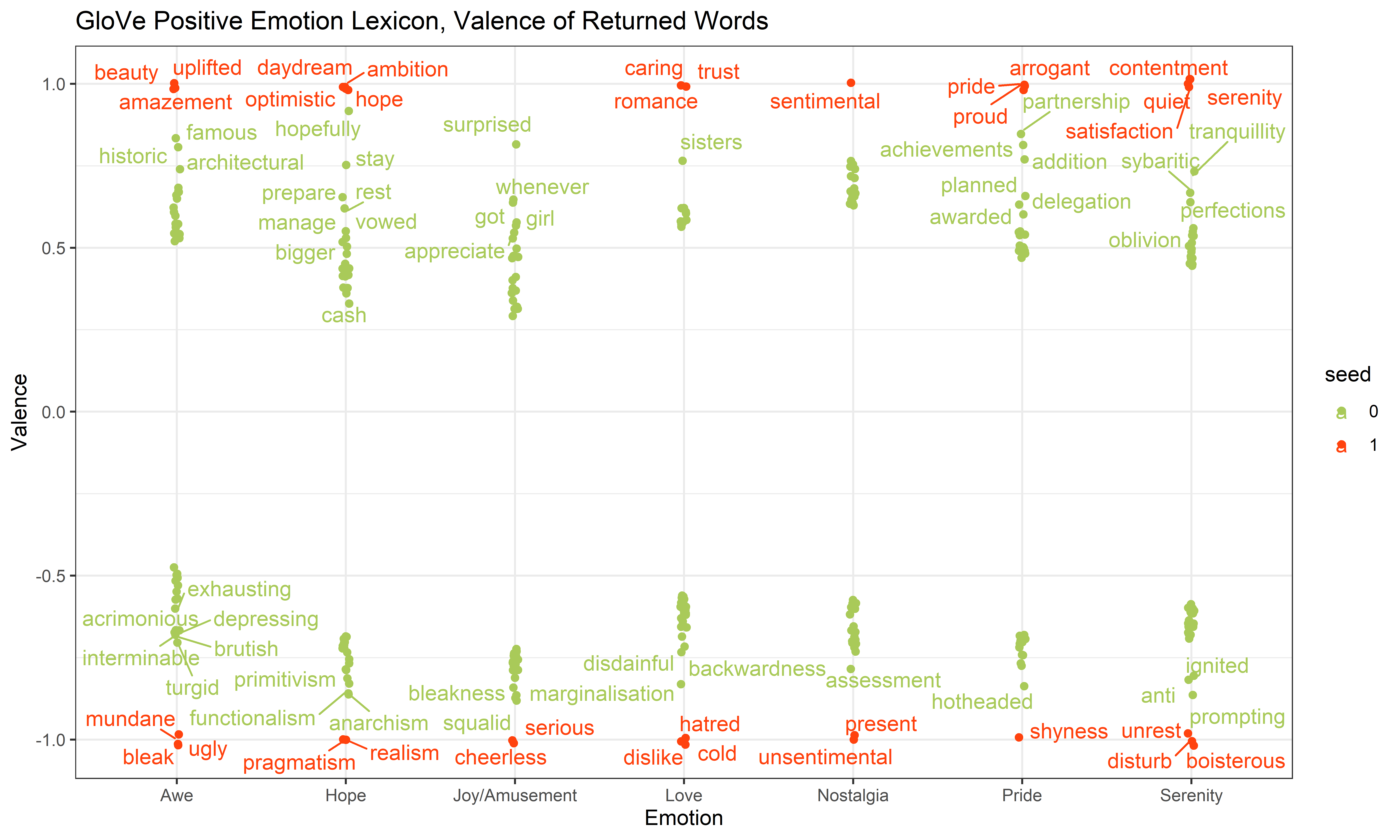}
\caption{Distinct positive emotion dictionary, GloVe}
\end{figure}

\hfill\break

\hfill\break

\hfill\break

\hfill\break

\hfill\break

\hfill\break

\hfill\break

\hfill\break

\hfill\break

\hfill\break

\hfill\break

\hfill\break

\hfill\break

\hfill\break

\newpage

\subsection{Differences between embedding
corpuses}\label{differences-between-embedding-corpuses}

An inspection of the most frequently appearing words in the FastText
Common Crawl and GloVe 6B corpuses shows the greater predominance of
formal, political and news-related terms in the GloVe data. As
elsewhere, the GloVe vocabulary is assumed to follow a decreasing order
by word frequency rank (as FastText is; see Pennington, Socher, and
Manning (2014), p.7, for a suggestion that this holds true for GloVe as
well). Word rank is converted into estimated usage counts using the
Mandelbrot formula,
\(count = corpus size \times \frac{1}{(rank + 2.7)^a}\) where corpus
size is set to 1 million tokens and \(a\) approximates 1, and is entered
as 1.

\begin{figure}
\centering
\includegraphics{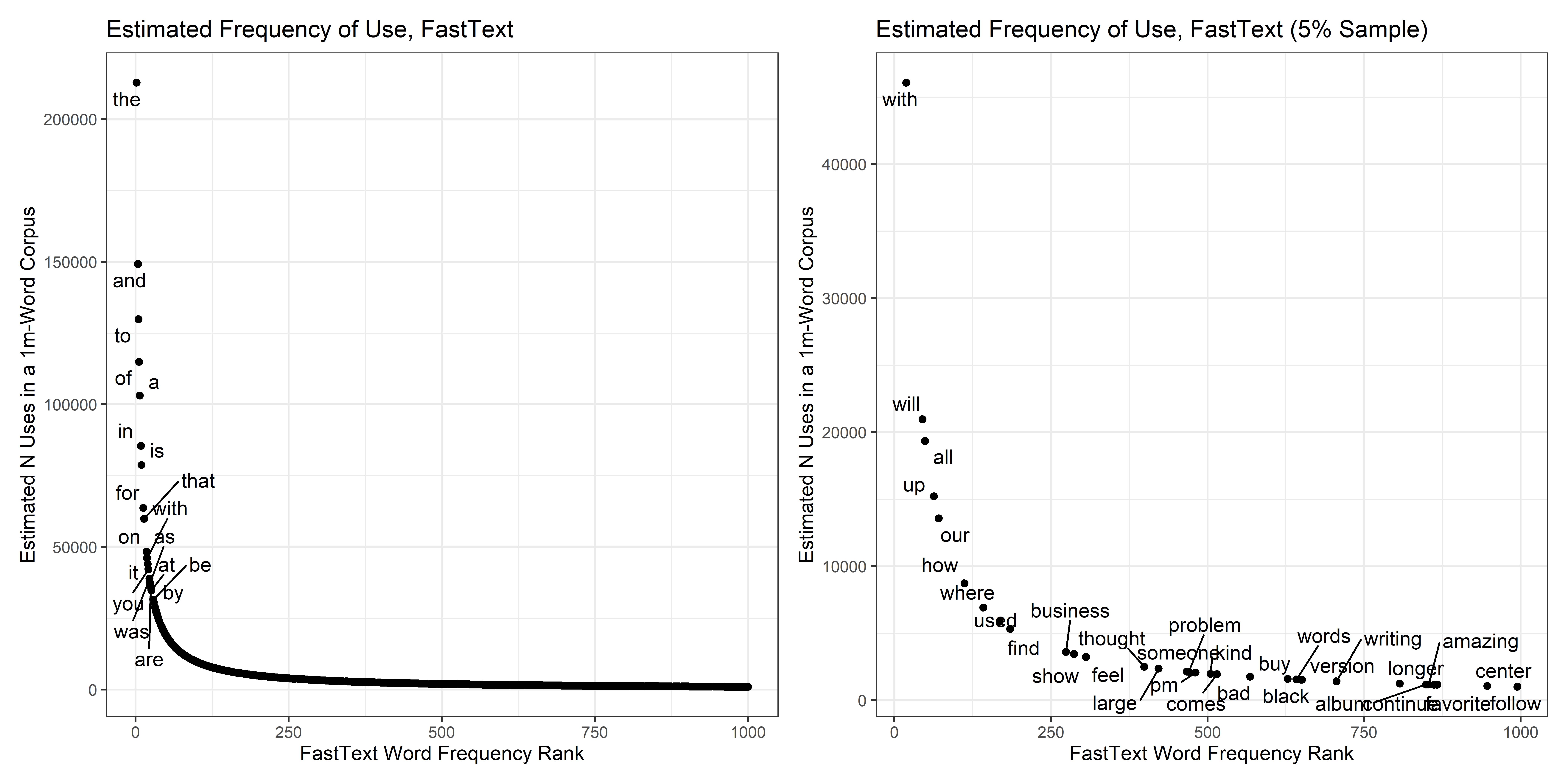}
\caption{Most frequent words in the Common Crawl FastText data}
\end{figure}

\begin{figure}
\centering
\includegraphics{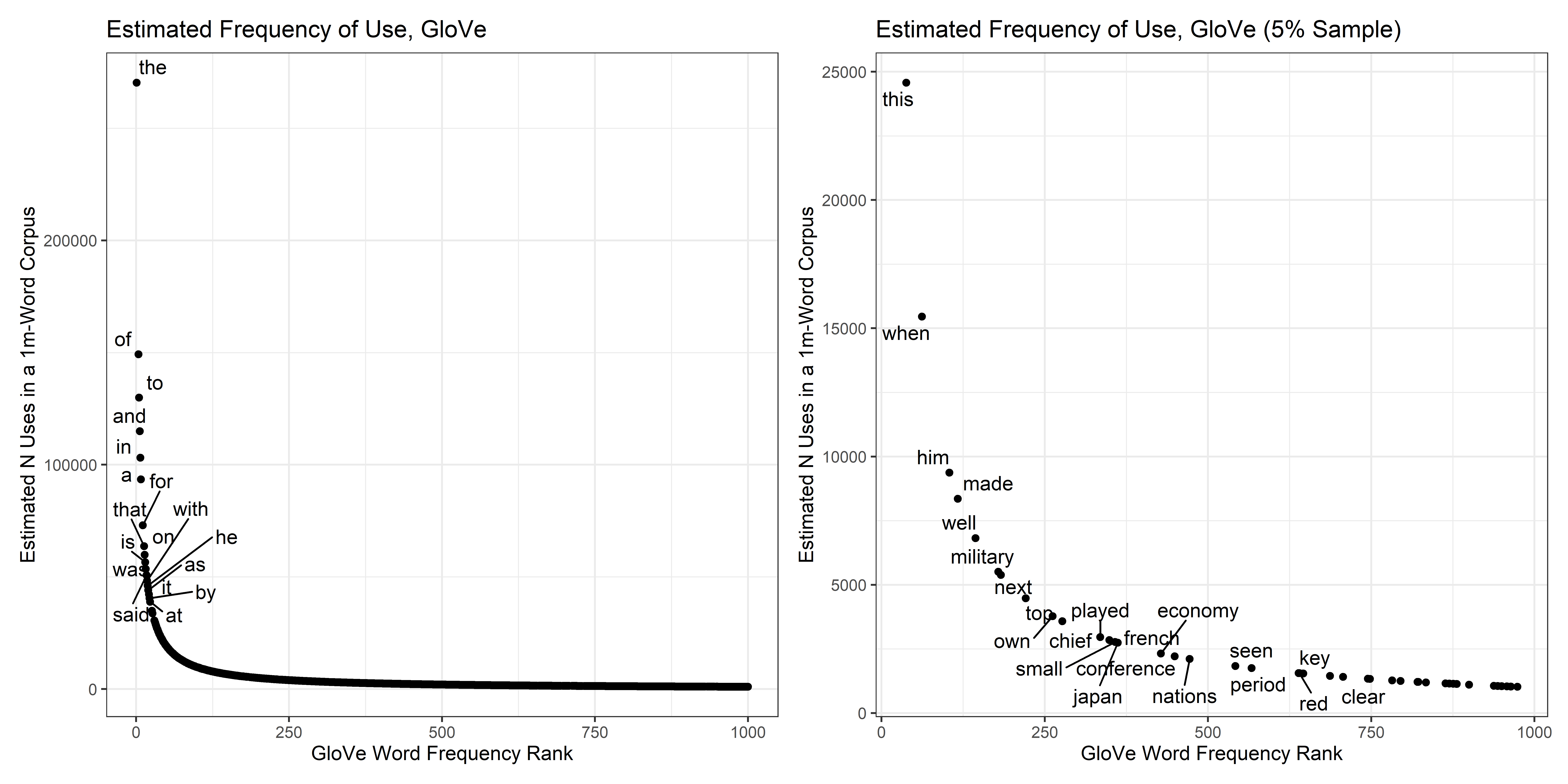}
\caption{Most frequent words in the Wikipedia 2014 + Gigaword 5 GloVe
data}
\end{figure}

\hfill\break

\hfill\break

\hfill\break

\hfill\break

\hfill\break

\hfill\break

\newpage

\subsection{Part 2: Accuracy Metrics}\label{part-2-accuracy-metrics}

Additional accuracy tests are shown below. Primarily, these apply
`valence scoring' of the dictionaries whereas the main-text tests
typically apply polarity scoring.

\paragraph{Paper Reviews (Conference Papers
data)}\label{paper-reviews-conference-papers-data}

\begin{longtable}[]{@{}
  >{\raggedright\arraybackslash}p{(\columnwidth - 6\tabcolsep) * \real{0.1481}}
  >{\raggedright\arraybackslash}p{(\columnwidth - 6\tabcolsep) * \real{0.1728}}
  >{\raggedright\arraybackslash}p{(\columnwidth - 6\tabcolsep) * \real{0.1728}}
  >{\raggedright\arraybackslash}p{(\columnwidth - 6\tabcolsep) * \real{0.5062}}@{}}
\caption{Performance of FastText and GloVe Sentiment Lexicons on
Conference Papers Dataset}\tabularnewline
\toprule\noalign{}
\begin{minipage}[b]{\linewidth}\raggedright
Lexicon
\end{minipage} & \begin{minipage}[b]{\linewidth}\raggedright
AdjRSquared
\end{minipage} & \begin{minipage}[b]{\linewidth}\raggedright
RMSE
\end{minipage} & \begin{minipage}[b]{\linewidth}\raggedright
Test
\end{minipage} \\
\midrule\noalign{}
\endfirsthead
\toprule\noalign{}
\begin{minipage}[b]{\linewidth}\raggedright
Lexicon
\end{minipage} & \begin{minipage}[b]{\linewidth}\raggedright
AdjRSquared
\end{minipage} & \begin{minipage}[b]{\linewidth}\raggedright
RMSE
\end{minipage} & \begin{minipage}[b]{\linewidth}\raggedright
Test
\end{minipage} \\
\midrule\noalign{}
\endhead
\bottomrule\noalign{}
\endlastfoot
GloVe & 0.04475 & 1.47116 & Conference Papers, Evaluators (Valence) \\
FastText & 0.040738 & 1.474246 & Conference Papers, Evaluators
(Valence) \\
PLOSlex & 0.035665 & 1.47814 & Conference Papers, Evaluators
(Valence) \\
NRC Emotion & -0.000411 & 1.505534 & Conference Papers, Evaluators
(Valence) \\
Lexicoder & not available & not available & Conference Papers,
Evaluators (Valence) \\
GloVe & 0.097052 & 0.975522 & Conference Papers, Researchers
(Valence) \\
FastText & 0.091427 & 0.978556 & Conference Papers, Researchers
(Valence) \\
PLOSlex & 0.085909 & 0.981523 & Conference Papers, Researchers
(Valence) \\
NRC Emotion & 0.005709 & 1.023676 & Conference Papers, Researchers
(Valence) \\
Lexicoder & not available & not available & Conference Papers,
Researchers (Valence) \\
\end{longtable}

\paragraph{Amazon Reviews,
alternatives}\label{amazon-reviews-alternatives}

Performance on three categories, polarity scored and disaggregated:

\begin{longtable}[]{@{}lrrl@{}}
\caption{Performance of FastText and GloVe Sentiment Lexicons on Amazon
Book Reviews (Linear Regression Models)}\tabularnewline
\toprule\noalign{}
Lexicon & AdjRSquared & RMSE & Test \\
\midrule\noalign{}
\endfirsthead
\toprule\noalign{}
Lexicon & AdjRSquared & RMSE & Test \\
\midrule\noalign{}
\endhead
\bottomrule\noalign{}
\endlastfoot
Lexicoder & 0.0042634 & 2.359200 & Amazon Book Reviews (n = 1000) \\
PLOSlex & 0.0037364 & 2.359824 & Amazon Book Reviews (n = 1000) \\
GloVe & 0.0028141 & 2.360916 & Amazon Book Reviews (n = 1000) \\
GPT 4o & 0.0028141 & 2.360916 & Amazon Book Reviews (n = 1000) \\
FastText & 0.0022108 & 2.361630 & Amazon Book Reviews (n = 1000) \\
NRC Emotion & -0.0001282 & 2.364396 & Amazon Book Reviews (n = 1000) \\
\end{longtable}

\begin{longtable}[]{@{}
  >{\raggedright\arraybackslash}p{(\columnwidth - 8\tabcolsep) * \real{0.0441}}
  >{\raggedright\arraybackslash}p{(\columnwidth - 8\tabcolsep) * \real{0.1765}}
  >{\raggedleft\arraybackslash}p{(\columnwidth - 8\tabcolsep) * \real{0.1765}}
  >{\raggedleft\arraybackslash}p{(\columnwidth - 8\tabcolsep) * \real{0.1324}}
  >{\raggedright\arraybackslash}p{(\columnwidth - 8\tabcolsep) * \real{0.4706}}@{}}
\caption{Performance of FastText and GloVe Sentiment Lexicons on Amazon
Music Reviews (Linear Regression Models)}\tabularnewline
\toprule\noalign{}
\begin{minipage}[b]{\linewidth}\raggedright
\end{minipage} & \begin{minipage}[b]{\linewidth}\raggedright
Lexicon
\end{minipage} & \begin{minipage}[b]{\linewidth}\raggedleft
AdjRSquared
\end{minipage} & \begin{minipage}[b]{\linewidth}\raggedleft
RMSE
\end{minipage} & \begin{minipage}[b]{\linewidth}\raggedright
Test
\end{minipage} \\
\midrule\noalign{}
\endfirsthead
\toprule\noalign{}
\begin{minipage}[b]{\linewidth}\raggedright
\end{minipage} & \begin{minipage}[b]{\linewidth}\raggedright
Lexicon
\end{minipage} & \begin{minipage}[b]{\linewidth}\raggedleft
AdjRSquared
\end{minipage} & \begin{minipage}[b]{\linewidth}\raggedleft
RMSE
\end{minipage} & \begin{minipage}[b]{\linewidth}\raggedright
Test
\end{minipage} \\
\midrule\noalign{}
\endhead
\bottomrule\noalign{}
\endlastfoot
7 & Lexicoder & 0.0042634 & 2.359200 & Amazon Music Reviews (n =
1000) \\
8 & PLOSlex & 0.0037364 & 2.359824 & Amazon Music Reviews (n = 1000) \\
9 & GloVe & 0.0028141 & 2.360916 & Amazon Music Reviews (n = 1000) \\
10 & FastText & 0.0022108 & 2.361630 & Amazon Music Reviews (n =
1000) \\
11 & GPT 4o & 0.0009791 & 2.363087 & Amazon Music Reviews (n = 1000) \\
12 & NRC Emotion & -0.0001282 & 2.364396 & Amazon Music Reviews (n =
1000) \\
\end{longtable}

\begin{longtable}[]{@{}
  >{\raggedright\arraybackslash}p{(\columnwidth - 8\tabcolsep) * \real{0.0423}}
  >{\raggedright\arraybackslash}p{(\columnwidth - 8\tabcolsep) * \real{0.1690}}
  >{\raggedleft\arraybackslash}p{(\columnwidth - 8\tabcolsep) * \real{0.1690}}
  >{\raggedleft\arraybackslash}p{(\columnwidth - 8\tabcolsep) * \real{0.1268}}
  >{\raggedright\arraybackslash}p{(\columnwidth - 8\tabcolsep) * \real{0.4930}}@{}}
\caption{Performance of FastText and GloVe Sentiment Lexicons on Amazon
Magazine Reviews (Linear Regression Models)}\tabularnewline
\toprule\noalign{}
\begin{minipage}[b]{\linewidth}\raggedright
\end{minipage} & \begin{minipage}[b]{\linewidth}\raggedright
Lexicon
\end{minipage} & \begin{minipage}[b]{\linewidth}\raggedleft
AdjRSquared
\end{minipage} & \begin{minipage}[b]{\linewidth}\raggedleft
RMSE
\end{minipage} & \begin{minipage}[b]{\linewidth}\raggedright
Test
\end{minipage} \\
\midrule\noalign{}
\endfirsthead
\toprule\noalign{}
\begin{minipage}[b]{\linewidth}\raggedright
\end{minipage} & \begin{minipage}[b]{\linewidth}\raggedright
Lexicon
\end{minipage} & \begin{minipage}[b]{\linewidth}\raggedleft
AdjRSquared
\end{minipage} & \begin{minipage}[b]{\linewidth}\raggedleft
RMSE
\end{minipage} & \begin{minipage}[b]{\linewidth}\raggedright
Test
\end{minipage} \\
\midrule\noalign{}
\endhead
\bottomrule\noalign{}
\endlastfoot
13 & Lexicoder & 0.0042634 & 2.359200 & Amazon Magazine Reviews (n =
1000) \\
14 & PLOSlex & 0.0037364 & 2.359824 & Amazon Magazine Reviews (n =
1000) \\
15 & GloVe & 0.0028141 & 2.360916 & Amazon Magazine Reviews (n =
1000) \\
16 & FastText & 0.0022108 & 2.361630 & Amazon Magazine Reviews (n =
1000) \\
17 & NRC Emotion & -0.0001282 & 2.364396 & Amazon Magazine Reviews (n =
1000) \\
18 & GPT 4o & -0.0008006 & 2.365191 & Amazon Magazine Reviews (n =
1000) \\
\end{longtable}

Performance on larger sample (n = 100,000) for Books data; performance
for full Digital Music (130.4k) and Magazine Subscriptions (71.5k) data:

\begin{longtable}[]{@{}lrrl@{}}
\caption{Performance of FastText and GloVe Sentiment Lexicons on Amazon
Reviews (Larger Samples)}\tabularnewline
\toprule\noalign{}
Lexicon & AdjRSquared & RMSE & Test \\
\midrule\noalign{}
\endfirsthead
\toprule\noalign{}
Lexicon & AdjRSquared & RMSE & Test \\
\midrule\noalign{}
\endhead
\bottomrule\noalign{}
\endlastfoot
Lexicoder & 0.0039617 & 2.354553 & Amazon Book Reviews \\
GloVe & 0.0038812 & 2.354648 & Amazon Book Reviews \\
FastText & 0.0037099 & 2.354851 & Amazon Book Reviews \\
PLOSlex & 0.0034618 & 2.355144 & Amazon Book Reviews \\
NRC Emotion & 0.0029539 & 2.355744 & Amazon Book Reviews \\
GloVe & 0.0445959 & 1.005045 & Amazon Music Reviews \\
FastText & 0.0441457 & 1.005282 & Amazon Music Reviews \\
Lexicoder & 0.0384453 & 1.008275 & Amazon Music Reviews \\
PLOSlex & 0.0355670 & 1.009783 & Amazon Music Reviews \\
NRC Emotion & 0.0353624 & 1.009890 & Amazon Music Reviews \\
GloVe & 0.1115829 & 1.377008 & Amazon Magazine Reviews \\
FastText & 0.1090344 & 1.378981 & Amazon Magazine Reviews \\
Lexicoder & 0.1044104 & 1.382555 & Amazon Magazine Reviews \\
PLOSlex & 0.0816985 & 1.399976 & Amazon Magazine Reviews \\
NRC Emotion & 0.0585990 & 1.417474 & Amazon Magazine Reviews \\
\end{longtable}

Amazon Books data, binarized and three-class rating DV models:

\begin{longtable}[]{@{}llrl@{}}
\caption{Performance of FastText and GloVe Sentiment Lexicons on Amazon
Book Reviews (Logistic Regression Models)}\tabularnewline
\toprule\noalign{}
Lexicon & Model & Accuracy & Test \\
\midrule\noalign{}
\endfirsthead
\toprule\noalign{}
Lexicon & Model & Accuracy & Test \\
\midrule\noalign{}
\endhead
\bottomrule\noalign{}
\endlastfoot
FastText & Logistic & 0.5499 & Amazon Book Reviews (Binary DV) \\
GloVe & Logistic & 0.5478 & Amazon Book Reviews (Binary DV) \\
Lexicoder & Logistic & 0.5454 & Amazon Book Reviews (Binary DV) \\
PLOSlex & Logistic & 0.5411 & Amazon Book Reviews (Binary DV) \\
NRC Emotion & Logistic & 0.5344 & Amazon Book Reviews (Binary DV) \\
FastText & Logistic & 0.4433 & Amazon Book Reviews (3-Class DV) \\
GloVe & Logistic & 0.4424 & Amazon Book Reviews (3-Class DV) \\
PLOSlex & Logistic & 0.4415 & Amazon Book Reviews (3-Class DV) \\
Lexicoder & Logistic & 0.4415 & Amazon Book Reviews (3-Class DV) \\
NRC Emotion & Logistic & 0.4412 & Amazon Book Reviews (3-Class DV) \\
\end{longtable}

\paragraph{Moral Foundations tests,
alternatives}\label{moral-foundations-tests-alternatives}

Accuracy scores for the three Reddit forums, scored separately, follow.
The main tests shown include posts annotated as `Non-moral', which other
moral frames datasets do not include.

\begin{longtable}[]{@{}
  >{\raggedright\arraybackslash}p{(\columnwidth - 8\tabcolsep) * \real{0.0349}}
  >{\raggedright\arraybackslash}p{(\columnwidth - 8\tabcolsep) * \real{0.2209}}
  >{\raggedleft\arraybackslash}p{(\columnwidth - 8\tabcolsep) * \real{0.1047}}
  >{\raggedleft\arraybackslash}p{(\columnwidth - 8\tabcolsep) * \real{0.1047}}
  >{\raggedright\arraybackslash}p{(\columnwidth - 8\tabcolsep) * \real{0.5349}}@{}}
\caption{Performance of FastText and GloVe Moral Foundations Lexicons on
Reddit Forums, Everyday Scenarios}\tabularnewline
\toprule\noalign{}
\begin{minipage}[b]{\linewidth}\raggedright
\end{minipage} & \begin{minipage}[b]{\linewidth}\raggedright
Lexicon
\end{minipage} & \begin{minipage}[b]{\linewidth}\raggedleft
F1
\end{minipage} & \begin{minipage}[b]{\linewidth}\raggedleft
Accuracy
\end{minipage} & \begin{minipage}[b]{\linewidth}\raggedright
Test
\end{minipage} \\
\midrule\noalign{}
\endfirsthead
\toprule\noalign{}
\begin{minipage}[b]{\linewidth}\raggedright
\end{minipage} & \begin{minipage}[b]{\linewidth}\raggedright
Lexicon
\end{minipage} & \begin{minipage}[b]{\linewidth}\raggedleft
F1
\end{minipage} & \begin{minipage}[b]{\linewidth}\raggedleft
Accuracy
\end{minipage} & \begin{minipage}[b]{\linewidth}\raggedright
Test
\end{minipage} \\
\midrule\noalign{}
\endhead
\bottomrule\noalign{}
\endlastfoot
25 & GPT 4o & 0.289361 & 0.289361 & Moral Frame Identification (Reddit,
Everyday) \\
26 & MFD & 0.293125 & 0.293125 & Moral Frame Identification (Reddit,
Everyday) \\
27 & MoralBERT & 0.344017 & 0.344017 & Moral Frame Identification
(Reddit, Everyday) \\
28 & FastText & 0.164755 & 0.164755 & Moral Frame Identification
(Reddit, Everyday) \\
29 & EMFD & 0.113802 & 0.113802 & Moral Frame Identification (Reddit,
Everyday) \\
30 & GloVe & 0.084907 & 0.084907 & Moral Frame Identification (Reddit,
Everyday) \\
31 & MoralStrength v1.1 & 0.287799 & 0.287799 & Moral Frame
Identification (Reddit, Everyday) \\
32 & MoralStrength v1 & 0.284894 & 0.284894 & Moral Frame Identification
(Reddit, Everyday) \\
\end{longtable}

\begin{longtable}[]{@{}
  >{\raggedright\arraybackslash}p{(\columnwidth - 8\tabcolsep) * \real{0.0323}}
  >{\raggedright\arraybackslash}p{(\columnwidth - 8\tabcolsep) * \real{0.2043}}
  >{\raggedleft\arraybackslash}p{(\columnwidth - 8\tabcolsep) * \real{0.0968}}
  >{\raggedleft\arraybackslash}p{(\columnwidth - 8\tabcolsep) * \real{0.0968}}
  >{\raggedright\arraybackslash}p{(\columnwidth - 8\tabcolsep) * \real{0.5699}}@{}}
\caption{Performance of FastText and GloVe Moral Foundations Lexicons on
Reddit Forums, French Politics}\tabularnewline
\toprule\noalign{}
\begin{minipage}[b]{\linewidth}\raggedright
\end{minipage} & \begin{minipage}[b]{\linewidth}\raggedright
Lexicon
\end{minipage} & \begin{minipage}[b]{\linewidth}\raggedleft
F1
\end{minipage} & \begin{minipage}[b]{\linewidth}\raggedleft
Accuracy
\end{minipage} & \begin{minipage}[b]{\linewidth}\raggedright
Test
\end{minipage} \\
\midrule\noalign{}
\endfirsthead
\toprule\noalign{}
\begin{minipage}[b]{\linewidth}\raggedright
\end{minipage} & \begin{minipage}[b]{\linewidth}\raggedright
Lexicon
\end{minipage} & \begin{minipage}[b]{\linewidth}\raggedleft
F1
\end{minipage} & \begin{minipage}[b]{\linewidth}\raggedleft
Accuracy
\end{minipage} & \begin{minipage}[b]{\linewidth}\raggedright
Test
\end{minipage} \\
\midrule\noalign{}
\endhead
\bottomrule\noalign{}
\endlastfoot
33 & GPT 4o & 0.240574 & 0.240574 & Moral Frame Identification (Reddit,
French Politics) \\
34 & MFD & 0.230494 & 0.230494 & Moral Frame Identification (Reddit,
French Politics) \\
35 & FastText & 0.109733 & 0.109733 & Moral Frame Identification
(Reddit, French Politics) \\
36 & MoralBERT & 0.192496 & 0.192496 & Moral Frame Identification
(Reddit, French Politics) \\
37 & MoralStrength v1.1 & 0.202723 & 0.202723 & Moral Frame
Identification (Reddit, French Politics) \\
38 & GloVe & 0.064263 & 0.064263 & Moral Frame Identification (Reddit,
French Politics) \\
39 & MoralStrength v1 & 0.213550 & 0.213550 & Moral Frame Identification
(Reddit, French Politics) \\
40 & EMFD & 0.066139 & 0.066139 & Moral Frame Identification (Reddit,
French Politics) \\
\end{longtable}

\begin{longtable}[]{@{}
  >{\raggedright\arraybackslash}p{(\columnwidth - 8\tabcolsep) * \real{0.0330}}
  >{\raggedright\arraybackslash}p{(\columnwidth - 8\tabcolsep) * \real{0.2088}}
  >{\raggedleft\arraybackslash}p{(\columnwidth - 8\tabcolsep) * \real{0.0989}}
  >{\raggedleft\arraybackslash}p{(\columnwidth - 8\tabcolsep) * \real{0.0989}}
  >{\raggedright\arraybackslash}p{(\columnwidth - 8\tabcolsep) * \real{0.5604}}@{}}
\caption{Performance of FastText and GloVe Moral Foundations Lexicons on
Reddit Forums, U.S. Politics}\tabularnewline
\toprule\noalign{}
\begin{minipage}[b]{\linewidth}\raggedright
\end{minipage} & \begin{minipage}[b]{\linewidth}\raggedright
Lexicon
\end{minipage} & \begin{minipage}[b]{\linewidth}\raggedleft
F1
\end{minipage} & \begin{minipage}[b]{\linewidth}\raggedleft
Accuracy
\end{minipage} & \begin{minipage}[b]{\linewidth}\raggedright
Test
\end{minipage} \\
\midrule\noalign{}
\endfirsthead
\toprule\noalign{}
\begin{minipage}[b]{\linewidth}\raggedright
\end{minipage} & \begin{minipage}[b]{\linewidth}\raggedright
Lexicon
\end{minipage} & \begin{minipage}[b]{\linewidth}\raggedleft
F1
\end{minipage} & \begin{minipage}[b]{\linewidth}\raggedleft
Accuracy
\end{minipage} & \begin{minipage}[b]{\linewidth}\raggedright
Test
\end{minipage} \\
\midrule\noalign{}
\endhead
\bottomrule\noalign{}
\endlastfoot
41 & GPT 4o & 0.271526 & 0.271526 & Moral Frame Identification (Reddit,
U.S. Politics) \\
42 & MFD & 0.257278 & 0.257278 & Moral Frame Identification (Reddit,
U.S. Politics) \\
43 & MoralBERT & 0.284207 & 0.284207 & Moral Frame Identification
(Reddit, U.S. Politics) \\
44 & FastText & 0.138314 & 0.138314 & Moral Frame Identification
(Reddit, U.S. Politics) \\
45 & EMFD & 0.107712 & 0.107712 & Moral Frame Identification (Reddit,
U.S. Politics) \\
46 & GloVe & 0.101538 & 0.101538 & Moral Frame Identification (Reddit,
U.S. Politics) \\
47 & MoralStrength v1.1 & 0.265703 & 0.265703 & Moral Frame
Identification (Reddit, U.S. Politics) \\
48 & MoralStrength v1 & 0.262838 & 0.262838 & Moral Frame Identification
(Reddit, U.S. Politics) \\
\end{longtable}

Alternatively, once `non-moral' entries are removed, performance is as
follows:

\begin{longtable}[]{@{}
  >{\raggedright\arraybackslash}p{(\columnwidth - 8\tabcolsep) * \real{0.0280}}
  >{\raggedright\arraybackslash}p{(\columnwidth - 8\tabcolsep) * \real{0.1776}}
  >{\raggedleft\arraybackslash}p{(\columnwidth - 8\tabcolsep) * \real{0.0841}}
  >{\raggedleft\arraybackslash}p{(\columnwidth - 8\tabcolsep) * \real{0.0841}}
  >{\raggedright\arraybackslash}p{(\columnwidth - 8\tabcolsep) * \real{0.6262}}@{}}
\caption{Performance of FastText and GloVe Moral Foundations Lexicons on
Reddit Forums, Everyday Scenarios}\tabularnewline
\toprule\noalign{}
\begin{minipage}[b]{\linewidth}\raggedright
\end{minipage} & \begin{minipage}[b]{\linewidth}\raggedright
Lexicon
\end{minipage} & \begin{minipage}[b]{\linewidth}\raggedleft
F1
\end{minipage} & \begin{minipage}[b]{\linewidth}\raggedleft
Accuracy
\end{minipage} & \begin{minipage}[b]{\linewidth}\raggedright
Test
\end{minipage} \\
\midrule\noalign{}
\endfirsthead
\toprule\noalign{}
\begin{minipage}[b]{\linewidth}\raggedright
\end{minipage} & \begin{minipage}[b]{\linewidth}\raggedright
Lexicon
\end{minipage} & \begin{minipage}[b]{\linewidth}\raggedleft
F1
\end{minipage} & \begin{minipage}[b]{\linewidth}\raggedleft
Accuracy
\end{minipage} & \begin{minipage}[b]{\linewidth}\raggedright
Test
\end{minipage} \\
\midrule\noalign{}
\endhead
\bottomrule\noalign{}
\endlastfoot
49 & MoralBERT & 0.668759 & 0.668759 & Moral Frame Identification
(Reddit, Everyday, 5 Moral Frames Only) \\
50 & EMFD & 0.245920 & 0.245920 & Moral Frame Identification (Reddit,
Everyday, 5 Moral Frames Only) \\
51 & FastText & 0.264479 & 0.264479 & Moral Frame Identification
(Reddit, Everyday, 5 Moral Frames Only) \\
52 & MFD & 0.299512 & 0.299512 & Moral Frame Identification (Reddit,
Everyday, 5 Moral Frames Only) \\
53 & GloVe & 0.188939 & 0.188939 & Moral Frame Identification (Reddit,
Everyday, 5 Moral Frames Only) \\
54 & GPT 4o & 0.208976 & 0.208976 & Moral Frame Identification (Reddit,
Everyday, 5 Moral Frames Only) \\
55 & MoralStrength v1.1 & 0.271075 & 0.271075 & Moral Frame
Identification (Reddit, Everyday, 5 Moral Frames Only) \\
56 & MoralStrength v1 & 0.235392 & 0.235392 & Moral Frame Identification
(Reddit, Everyday, 5 Moral Frames Only) \\
\end{longtable}

\begin{longtable}[]{@{}
  >{\raggedright\arraybackslash}p{(\columnwidth - 8\tabcolsep) * \real{0.0263}}
  >{\raggedright\arraybackslash}p{(\columnwidth - 8\tabcolsep) * \real{0.1667}}
  >{\raggedleft\arraybackslash}p{(\columnwidth - 8\tabcolsep) * \real{0.0789}}
  >{\raggedleft\arraybackslash}p{(\columnwidth - 8\tabcolsep) * \real{0.0789}}
  >{\raggedright\arraybackslash}p{(\columnwidth - 8\tabcolsep) * \real{0.6491}}@{}}
\caption{Performance of FastText and GloVe Moral Foundations Lexicons on
Reddit Forums, French Politics}\tabularnewline
\toprule\noalign{}
\begin{minipage}[b]{\linewidth}\raggedright
\end{minipage} & \begin{minipage}[b]{\linewidth}\raggedright
Lexicon
\end{minipage} & \begin{minipage}[b]{\linewidth}\raggedleft
F1
\end{minipage} & \begin{minipage}[b]{\linewidth}\raggedleft
Accuracy
\end{minipage} & \begin{minipage}[b]{\linewidth}\raggedright
Test
\end{minipage} \\
\midrule\noalign{}
\endfirsthead
\toprule\noalign{}
\begin{minipage}[b]{\linewidth}\raggedright
\end{minipage} & \begin{minipage}[b]{\linewidth}\raggedright
Lexicon
\end{minipage} & \begin{minipage}[b]{\linewidth}\raggedleft
F1
\end{minipage} & \begin{minipage}[b]{\linewidth}\raggedleft
Accuracy
\end{minipage} & \begin{minipage}[b]{\linewidth}\raggedright
Test
\end{minipage} \\
\midrule\noalign{}
\endhead
\bottomrule\noalign{}
\endlastfoot
57 & MoralBERT & 0.498856 & 0.498856 & Moral Frame Identification
(Reddit, French Politics, 5 Moral Frames Only) \\
58 & GloVe & 0.209098 & 0.209098 & Moral Frame Identification (Reddit,
French Politics, 5 Moral Frames Only) \\
59 & EMFD & 0.205624 & 0.205624 & Moral Frame Identification (Reddit,
French Politics, 5 Moral Frames Only) \\
60 & FastText & 0.224651 & 0.224651 & Moral Frame Identification
(Reddit, French Politics, 5 Moral Frames Only) \\
61 & MFD & 0.244569 & 0.244569 & Moral Frame Identification (Reddit,
French Politics, 5 Moral Frames Only) \\
62 & MoralStrength v1.1 & 0.218832 & 0.218832 & Moral Frame
Identification (Reddit, French Politics, 5 Moral Frames Only) \\
63 & GPT 4o & 0.178480 & 0.178480 & Moral Frame Identification (Reddit,
French Politics, 5 Moral Frames Only) \\
64 & MoralStrength v1 & 0.190650 & 0.190650 & Moral Frame Identification
(Reddit, French Politics, 5 Moral Frames Only) \\
\end{longtable}

\begin{longtable}[]{@{}
  >{\raggedright\arraybackslash}p{(\columnwidth - 8\tabcolsep) * \real{0.0268}}
  >{\raggedright\arraybackslash}p{(\columnwidth - 8\tabcolsep) * \real{0.1696}}
  >{\raggedleft\arraybackslash}p{(\columnwidth - 8\tabcolsep) * \real{0.0804}}
  >{\raggedleft\arraybackslash}p{(\columnwidth - 8\tabcolsep) * \real{0.0804}}
  >{\raggedright\arraybackslash}p{(\columnwidth - 8\tabcolsep) * \real{0.6429}}@{}}
\caption{Performance of FastText and GloVe Moral Foundations Lexicons on
Reddit Forums, U.S. Politics}\tabularnewline
\toprule\noalign{}
\begin{minipage}[b]{\linewidth}\raggedright
\end{minipage} & \begin{minipage}[b]{\linewidth}\raggedright
Lexicon
\end{minipage} & \begin{minipage}[b]{\linewidth}\raggedleft
F1
\end{minipage} & \begin{minipage}[b]{\linewidth}\raggedleft
Accuracy
\end{minipage} & \begin{minipage}[b]{\linewidth}\raggedright
Test
\end{minipage} \\
\midrule\noalign{}
\endfirsthead
\toprule\noalign{}
\begin{minipage}[b]{\linewidth}\raggedright
\end{minipage} & \begin{minipage}[b]{\linewidth}\raggedright
Lexicon
\end{minipage} & \begin{minipage}[b]{\linewidth}\raggedleft
F1
\end{minipage} & \begin{minipage}[b]{\linewidth}\raggedleft
Accuracy
\end{minipage} & \begin{minipage}[b]{\linewidth}\raggedright
Test
\end{minipage} \\
\midrule\noalign{}
\endhead
\bottomrule\noalign{}
\endlastfoot
65 & MoralBERT & 0.622706 & 0.622706 & Moral Frame Identification
(Reddit, U.S. Politics, 5 Moral Frames Only) \\
66 & GloVe & 0.242333 & 0.242333 & Moral Frame Identification (Reddit,
U.S. Politics, 5 Moral Frames Only) \\
67 & EMFD & 0.257059 & 0.257059 & Moral Frame Identification (Reddit,
U.S. Politics, 5 Moral Frames Only) \\
68 & MFD & 0.294243 & 0.294243 & Moral Frame Identification (Reddit,
U.S. Politics, 5 Moral Frames Only) \\
69 & FastText & 0.244534 & 0.244534 & Moral Frame Identification
(Reddit, U.S. Politics, 5 Moral Frames Only) \\
70 & GPT 4o & 0.203847 & 0.203847 & Moral Frame Identification (Reddit,
U.S. Politics, 5 Moral Frames Only) \\
71 & MoralStrength v1.1 & 0.292485 & 0.292485 & Moral Frame
Identification (Reddit, U.S. Politics, 5 Moral Frames Only) \\
72 & MoralStrength v1 & 0.242561 & 0.242561 & Moral Frame Identification
(Reddit, U.S. Politics, 5 Moral Frames Only) \\
\end{longtable}

\newpage

\subsection{Part 3: Performance on Unlabelled Data, and Similarities
between
Lexicons}\label{part-3-performance-on-unlabelled-data-and-similarities-between-lexicons}

\paragraph{Distinct Positive Emotions in U.S. Presidents' Inaugural
Speeches}\label{distinct-positive-emotions-in-u.s.-presidents-inaugural-speeches-1}

\(\qquad\) Entire speeches are scored by valence (and general sentiment)
with the FastText dictionary, below.

\begin{figure}
\centering
\includegraphics{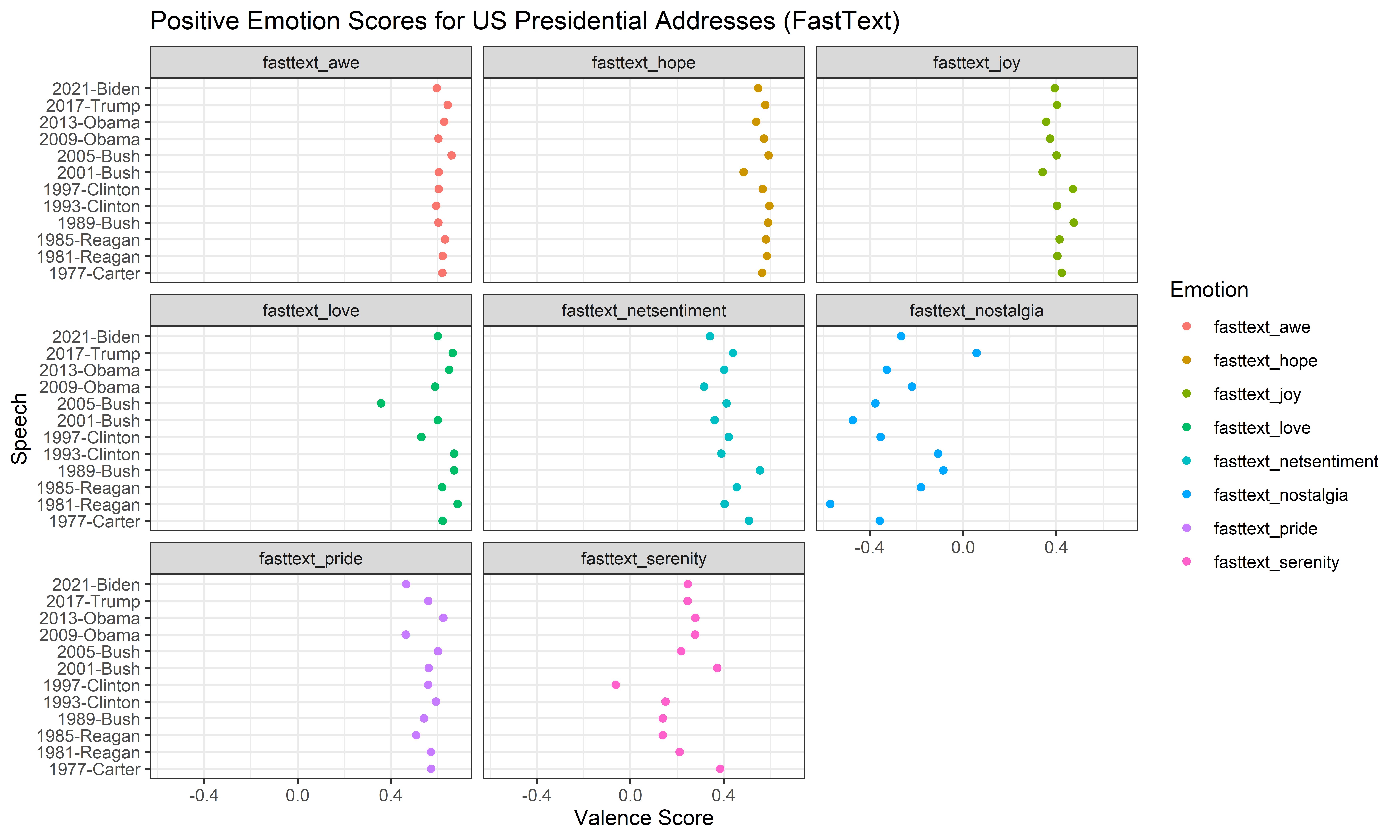}
\caption{U.S. inaugural addresses, scored for distinct positive
emotions}
\end{figure}

\hfill\break

\paragraph{Similarity between
lexicons}\label{similarity-between-lexicons}

\hfill\break

\(\qquad\) As in the main text, words with the highest sentiment valence
discrepancies with other sentiment dictionaries are plotted below. (Seed
words are excluded, as these are automatically given a valence of -1 or
1).

\(\qquad\) NRC emotion lexicon words are attributed valence scores using
the `intensities' attributed to them in a second NRC emotion dictionary,
the Emotion Intensity Lexicon (Saif M. Mohammad 2018). Emotion
intensities for negative emotions (fear, anger, disgust, sadness, and
the `anticipation' and `surprise' words labelled as Negative in the
Emotion Lexicon) are presented as negative scores.

\begin{figure}
\centering
\includegraphics{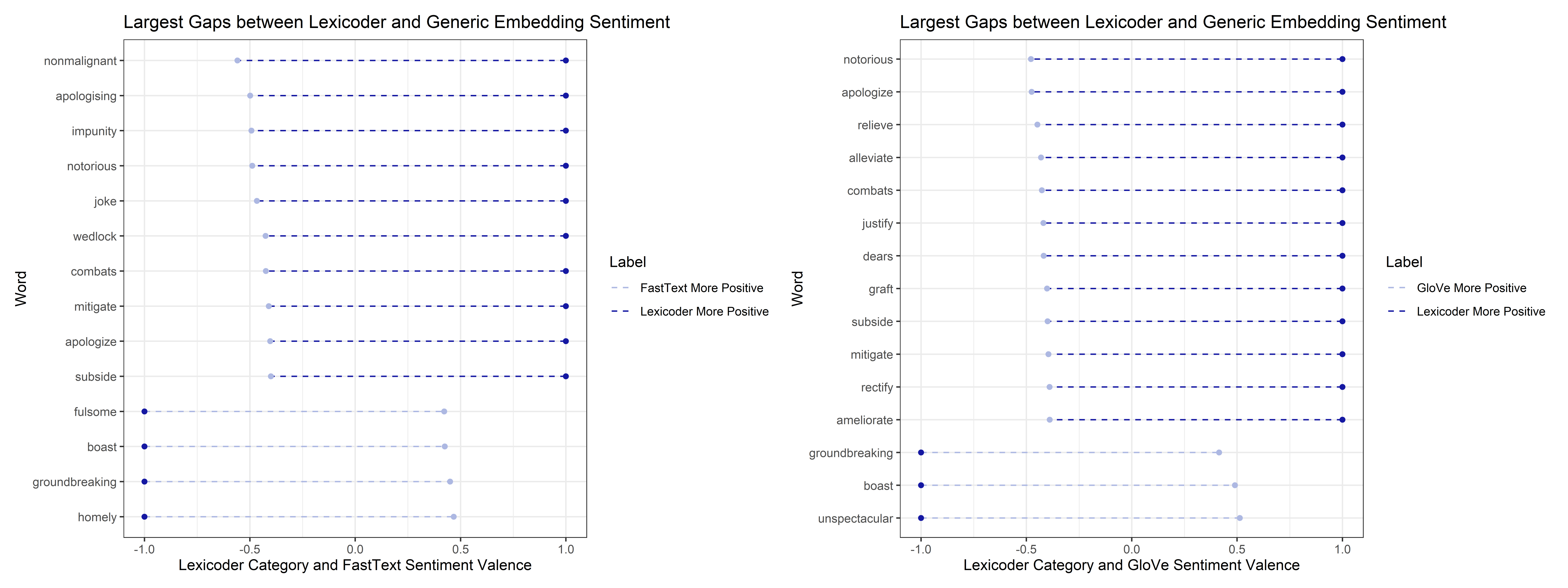}
\caption{Comparison of generic embedding dictionaries to Lexicoder
(crowdsourced)}
\end{figure}

\begin{figure}
\centering
\includegraphics{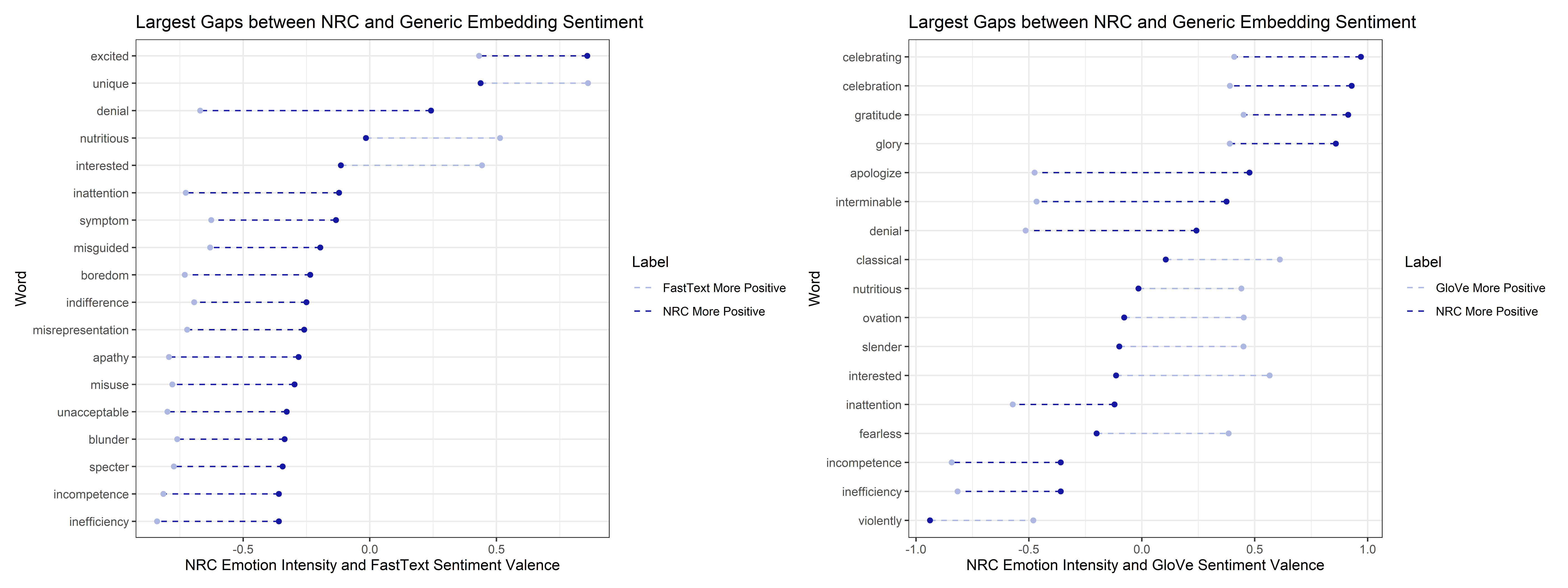}
\caption{Comparison of generic embedding dictionaries to NRC Emotion
Lexicons (crowdsourced)}
\end{figure}

\begin{figure}
\centering
\includegraphics{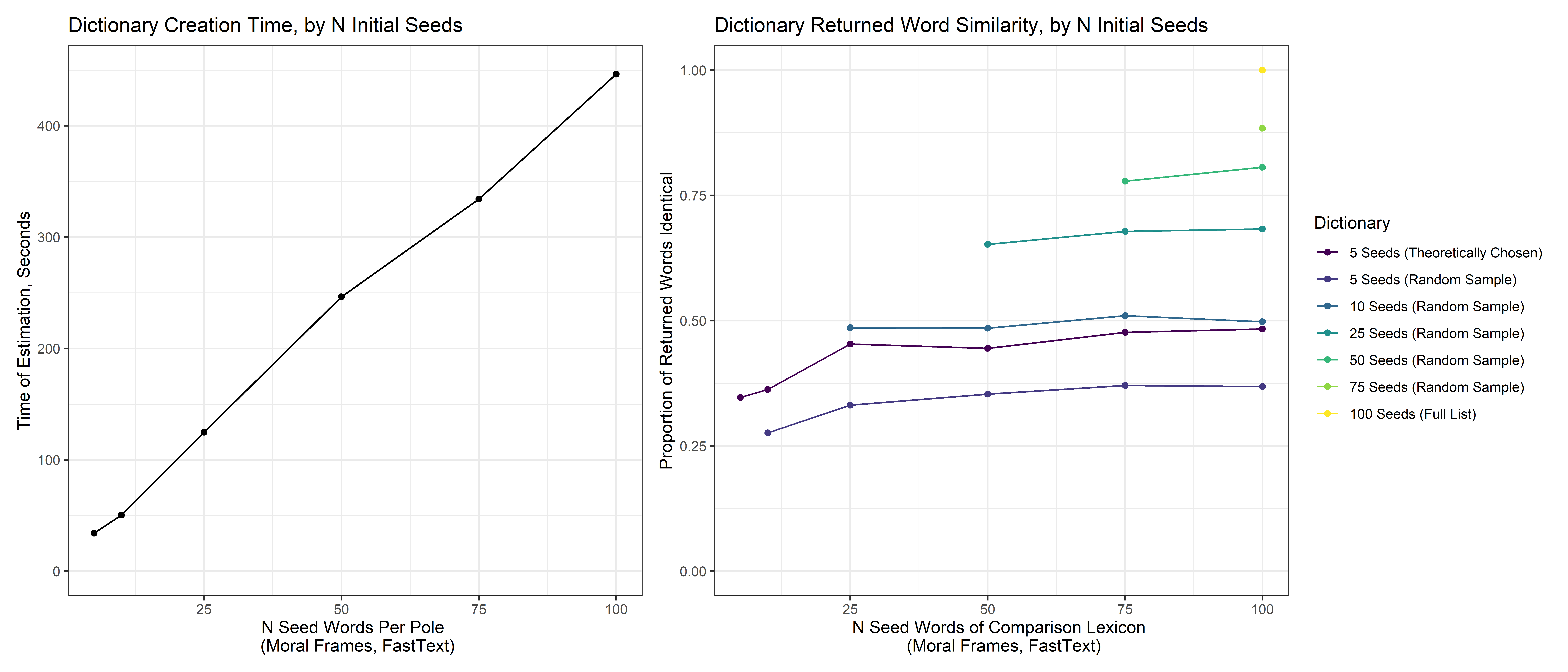}
\caption{Comparison of FastText lexicon composition by number of initial
seed words}
\end{figure}

\begin{figure}
\centering
\includegraphics{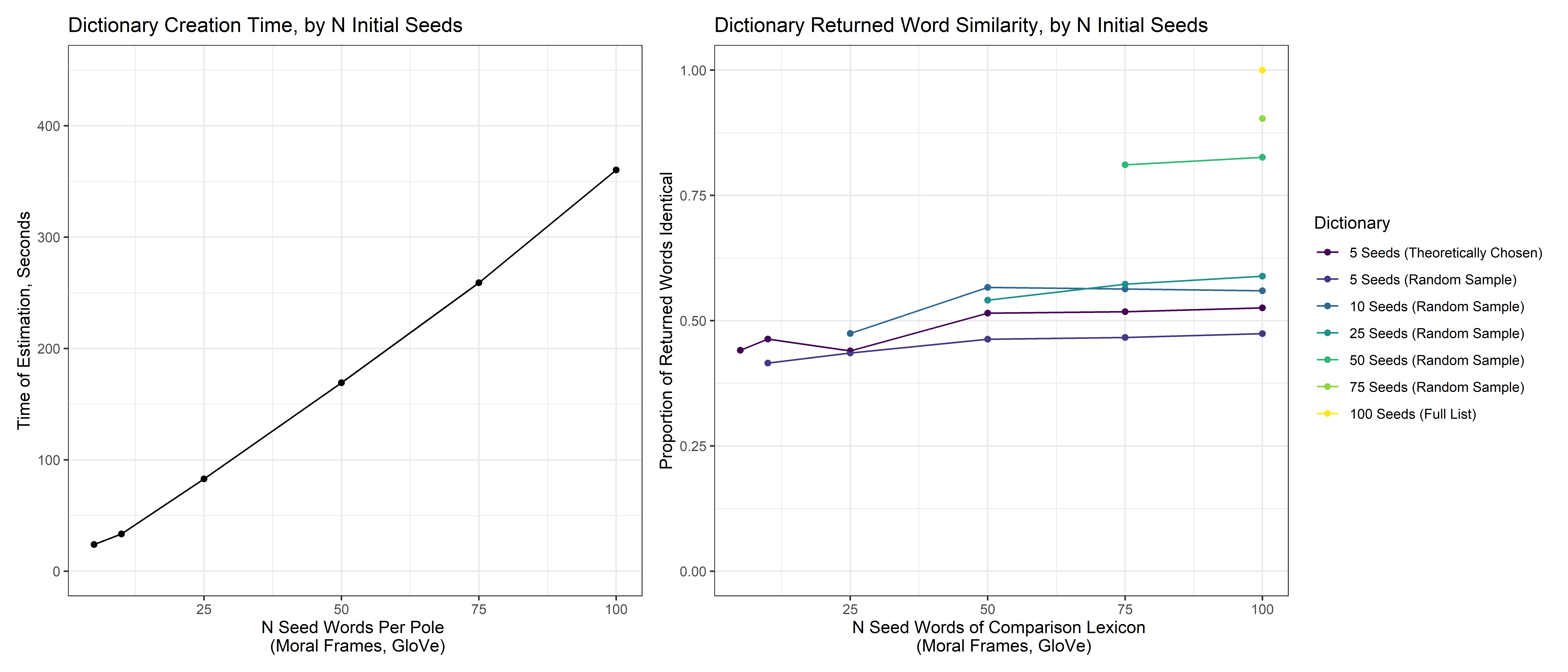}
\caption{Comparison of GloVe lexicon composition by number of initial
seed words}
\end{figure}

\end{document}